\documentclass{article}

\PassOptionsToPackage{numbers, compress}{natbib}
\usepackage[preprint]{neurips_2024}
\usepackage[T1]{fontenc}
\usepackage{amsmath,amssymb}
\usepackage{graphicx}
\usepackage{booktabs}
\usepackage{multirow}
\usepackage{xcolor}
\usepackage{caption}
\usepackage{listings}
\usepackage{float}
\usepackage[expansion=false]{microtype}
\usepackage[htt]{hyphenat}
\usepackage{xurl}
\usepackage{hyperref}
\DeclareRobustCommand{\codepath}[1]{\texttt{\detokenize{#1}}}
\usepackage{tikz}
\usetikzlibrary{positioning,shapes.geometric,fit,calc,backgrounds}
\hypersetup{colorlinks=true, linkcolor=blue, citecolor=blue, urlcolor=blue}

\makeatletter
\renewcommand{\@noticestring}{}
\makeatother

\title{Distilling Foundation Models for Agentic What-If Reasoning: \\
Cost, Latency, and Governance in a Hybrid LLM+SLM Architecture}

\author{%
  Sourish Dey\thanks{Code, data, and evaluation artefacts: \url{https://github.com/nitsourish/Model-Distillation-Agentic-Workflow}} \\
  Machine Learning \\
  SumUp, Berlin \\
  \texttt{sourish.dey@sumup.com} \\
  \texttt{sourish.syntel@gmail.com} \\
  \And
  Aditya Kumar \\
  Institute of Physics \\
  Johannes Gutenberg-Universit{\"a}t Mainz \\
  55128 Mainz, Germany \\
  \texttt{kuaditya@uni-mainz.de} \\
}
\begin{document}
\maketitle

\begin{abstract}
Tabular foundation models deliver strong zero-training predictive performance via in-context learning, but their high inference latency makes them impractical as hot-path decision backends in interactive agentic loops. We distill a TabPFN teacher into a compact feed-forward student across a business-decision simulation on UCI Adult and five OpenML benchmarks: the classification head compresses 53.2M parameters to 8{,}546 (6{,}220$\times$); the deployed two-head loan pipeline compresses 111.4M parameters to 17{,}059 (6{,}532$\times$). The student retains 95.4--100.5\% accuracy and 96.8--100.0\% AUC (Table~\ref{tab:results} point estimates from the same checkpoints as Table~\ref{tab:extended-metrics}; lowest accuracy retention is credit-g at 95.4\%), with an $\alpha=0$ hard-label control showing the teacher's soft targets provide a 2.1--7.0 AUC point gain. Embedding the distilled student in a hybrid LangGraph architecture (restricting the cloud LLM to tool selection and schema parsing while serving prediction and response formatting locally) reduces end-to-end latency by 3.8--4.8$\times$ across CPU and GPU profiles (22.91s vs.\ 87.95s on CPU; 1.53s vs.\ 5.86s on GPU; 4.96s vs.\ 23.94s on Apple M5 Pro). Span-level attribution shows that this wall-clock gain is almost entirely from removing in-context TabPFN from the hot path; fewer cloud round-trips cut token cost (2.1$\times$) rather than latency. Tool-call accuracy improves (0.98 vs.\ 0.90). Raw 14-column feature rows stay on-premises, but free-text queries and \texttt{customer\_id} still cross the cloud API. Cloud token cost drops 2.1$\times$, though hardware amortization requires multi-million-query scale. Both pipelines are fully instrumented with LangSmith tracing, and all artefacts are open-source.
\end{abstract}


\section{Introduction}

Foundation models have made few-shot, near-zero-training-cost prediction practical even for structured (tabular) data. Tabular foundation models such as TabPFN \citep{hollmann2023tabpfn,hollmann2025tabpfnv2,priorlabs2024} perform in-context learning: they ingest the entire training set as context at inference time and require no gradient-based fitting on the target dataset. This makes them attractive teachers for business-decision tasks where labelled data is scarce. However, their inference cost scales with context size; we measure single-call tabular foundation-model latencies of 12.6s (CPU) and 0.84s (GPU) per prediction (Section~\ref{sec:hybrid-results}), creating an unacceptable bottleneck inside interactive, multi-step agentic loops. 

Concurrently, the dominant deployment pattern delegates both tool orchestration and final response generation to a cloud LLM (e.g., GPT-4o-mini). In decision systems, this repeatedly transmits sensitive attributes (age, income, marital status) to external APIs, incurring unnecessary token latency and data-governance risks.

We evaluate whether \textbf{knowledge distillation of the tabular predictor}, combined with an asymmetric \textbf{hybrid orchestration split} that reserves the cloud LLM for tool selection and parsing while hosting prediction and response generation on-premises, can retain task quality while reducing latency, operational cost, and sensitive-data exposure. We make three contributions:
\begin{enumerate}
\item \textbf{TabPFN$\to$MLP distillation with $\alpha$ controls.} We compress a 53.2M-parameter TabPFN classifier into an 8{,}546-parameter student (6{,}220$\times$; 111.4M$\to$17{,}059 / 6{,}532$\times$ for the two-head loan pipeline) across UCI Adult and OpenML benchmarks, with hard-label-only ($\alpha=0$) controls showing a 2.1--7.0 AUC point gain from teacher soft targets (Sections~\ref{sec:distillation-results} and~\ref{sec:m1-control}).
\item \textbf{Hybrid agentic split, cost, and latency.} We evaluate a LangGraph pipeline against a full-LLM baseline on 50 what-if queries. End-to-end latency falls 3.8--4.8$\times$ (driven by removing TabPFN from the hot path; Section~\ref{sec:hybrid-results}), tool-call accuracy rises from 0.90 to 0.98, and cloud token cost falls 2.1$\times$.
\item \textbf{Governance and subgroup fairness.} We audit data minimization under GDPR and evaluate subgroup fairness, showing that aggregate retention can hide race equalized-odds gaps that widen under compression (Section~\ref{sec:discussion}).
\end{enumerate}
A recap of distillation versus fine-tuning and the implemented KD losses appears in Section~\ref{sec:theory} as background, not as a theoretical contribution. LangSmith tracing (Section~\ref{sec:observability}) is an implementation detail of the evaluated pipelines.

\section{Related Work}

\textbf{Knowledge distillation.} \citet{hinton2015distilling} introduced the canonical teacher-student framework matching softened output distributions alongside hard labels, building on model-compression foundations \citep{caruana2006model,ba2014deep,gou2021kdsurvey}. While widely applied to vision and language encoders (e.g., DistilBERT \citep{sanh2019distilbert}, TinyBERT \citep{jiao2020tinybert}), tabular foundation model distillation is less explored. Recent efforts condense TabPFN context representations \citep{ma2024incontext,feuer2024tunetables}. Our work focuses on compressing a fitted TabPFN teacher into an independent, lightweight MLP, comparing against plain MLP baselines \citep{gorishniy2021revisiting} and evaluating subgroup compression bias \citep{hooker2020characterising}.

\textbf{Tabular foundation models and adaptation.} TabPFN \citep{hollmann2023tabpfn,hollmann2025tabpfnv2,priorlabs2024} reframes tabular learning as in-context transformer inference. In contrast to fine-tuning (ULMFiT \citep{howard2018ulmfit}, instruction tuning \citep{ouyang2022instructgpt}) or parameter-efficient adapters like LoRA \citep{hu2022lora} which preserve model architecture and inference compute, distillation drastically alters the operational footprint.

\textbf{Agentic workflows and evaluation.} ReAct \citep{yao2023react}, Toolformer \citep{schick2023toolformer}, and frameworks like AutoGen \citep{wu2023autogen} and LangGraph \citep{langgraph2024} formalize multi-step tool orchestration. We assess small-model parsing capabilities zero-shot (Section~\ref{sec:slm-ablation}) and employ the RAGAS faithfulness/relevance triad \citep{es2024ragastriad} via an LLM judge.

\section{Background: Distillation versus Fine-Tuning}
\label{sec:theory}

This section recaps a standard distinction and records the losses used in our scripts. Fine-tuning and distillation both adapt learned knowledge to a target task, but they differ in what is optimized and what capacity constraint is imposed on the result.

\subsection{Fine-tuning}

Fine-tuning starts from a pretrained parameter vector $\theta_0$ of a fixed architecture $f_\theta$ and continues optimization on a target-task dataset $\mathcal{D}_{\text{tgt}} = \{(x_i, y_i)\}_{i=1}^N$:
\begin{equation}
\theta^\star = \arg\min_{\theta} \; \mathbb{E}_{(x,y)\sim \mathcal{D}_{\text{tgt}}} \Big[ \mathcal{L}_{\text{task}}\big(f_\theta(x), y\big) \Big], \qquad \theta \text{ initialized at } \theta_0.
\label{eq:finetune}
\end{equation}
In this setting, the adapted model retains the architecture of the pretrained checkpoint. In parameter-efficient variants such as LoRA \citep{hu2022lora}, weight updates are parameterized via low-rank decomposition $\Delta W = BA$ with $B \in \mathbb{R}^{d\times r}$, $A \in \mathbb{R}^{r\times k}$, $r \ll \min(d,k)$, leaving the original weights $W$ frozen. Fine-tuning adjusts the learned representations of a fixed-capacity architecture without substantially altering inference compute or memory requirements.

\subsection{Distillation}

Distillation trains a separate, compact student architecture $g_\phi$ to reproduce the behavior of a fixed teacher $f_{\theta_T}$ with frozen weights, using the teacher's output distribution as an additional supervisory signal alongside ground-truth labels:
\begin{equation}
\phi^\star = \arg\min_{\phi} \; \mathbb{E}_{x\sim \mathcal{D}} \Big[ \mathcal{L}_{\text{KD}}\big(g_\phi(x), f_{\theta_T}(x), y\big) \Big].
\label{eq:distill}
\end{equation}
Because $g_\phi$ and $f_{\theta_T}$ need not share architectural families or internal representations, distillation enables replacing a $\sim$53--111M-parameter transformer with an 8.5K-parameter feed-forward network (Section~\ref{sec:distillation-results}), achieving compression that architectural preservation during fine-tuning precludes.

\paragraph{Classification loss (as implemented, \texttt{run\_distillation.py}).} For classification we use the canonical \citet{hinton2015distilling} soft-target objective. Let $z_S, z_T \in \mathbb{R}^{C}$ be student and teacher logits, $T>0$ a softening temperature, $y$ the hard label, and $\alpha \in [0,1]$ the soft/hard mixing weight:
\begin{align}
\mathcal{L}_{\text{soft}} &= T^2 \cdot D_{\text{KL}}\Big( \operatorname{softmax}(z_T / T) \,\Big\|\, \operatorname{softmax}(z_S / T) \Big) \\
\mathcal{L}_{\text{hard}} &= \operatorname{CrossEntropy}(z_S, y) \\
\mathcal{L}_{\text{cls}}(z_S, z_T, y) &= \alpha \, \mathcal{L}_{\text{soft}} + (1-\alpha)\, \mathcal{L}_{\text{hard}}
\label{eq:kd-cls}
\end{align}
In our experiments, $T=2.0$ and $\alpha=0.7$, training the student primarily to match the teacher's softened class probabilities (which encode inter-class confidence structures in addition to the arg-max prediction) and secondarily to match ground-truth labels. The $T^2$ multiplier compensates for gradient magnitude scaling from logit division by $T$ \citep{hinton2015distilling}.

\paragraph{Regression loss (as implemented).} For the continuous regression head there is no softmax to soften, so the analogous soft/hard split uses mean-squared error against the teacher's point prediction $\hat{y}_T = f_{\theta_T}(x)$ and against the true label $y$:
\begin{equation}
\mathcal{L}_{\text{reg}}(\hat y_S, \hat y_T, y) = \alpha \cdot \operatorname{MSE}(\hat y_S, \hat y_T) + (1-\alpha)\cdot \operatorname{MSE}(\hat y_S, y), \qquad \alpha = 0.7.
\label{eq:kd-reg}
\end{equation}

\begin{table}[tbp]
\centering
\footnotesize
\resizebox{\textwidth}{!}{%
\begin{tabular}{p{2.6cm}p{6.4cm}p{6.4cm}}
\toprule
\textbf{Dimension} & \textbf{Fine-tuning} & \textbf{Distillation} \\
\midrule
Primary objective & Specialize general model to narrower task distribution & Compress large model into smaller, faster architecture \\
Target architecture & Unchanged (full FT) or near-identical (LoRA, adapters) & Student architecture unconstrained; typically $\ll$ teacher \\
Inference cost & Unchanged ($\approx$ same latency, memory, compute) & Substantially reduced ($\propto$ student parameter count) \\
Label requirements & Requires task-specific ground-truth labels $\mathcal{D}$ & Can use soft targets from unlabelled transfer sets \\
Representational fidelity & Preserves internal representations of the base model & Transfers functional input-output mapping; internals differ \\
Cold-start latency & High (must load large model into memory) & Low (compact student fits in lightweight environments) \\
Production role & Not deployed (explored zero-shot in Section~\ref{sec:slm-ablation}) & Core tabular prediction engine (Sections~\ref{sec:distillation-results} and \ref{sec:system}) \\
\bottomrule
\end{tabular}%
}
\caption{Fine-tuning vs.\ distillation: architectural and operational dimensions.}
\label{tab:finetune-vs-distill}
\end{table}

\section{System Architecture}
\label{sec:system}

The two agentic pipelines evaluated end-to-end in Section~\ref{sec:hybrid-results} (see repository \codepath{scripts/hybrid_architecture.py}) are as follows.

\begin{itemize}
\item \textbf{Full-LLM baseline (\texttt{FullLLMAgent}).} A single cloud model (\texttt{gpt-4o-mini}) receives the system prompt, the user query, and the tool schemas; it decides which tool to call, extracts parameters (including free-text what-if modifications, e.g. \emph{``what if customer 123 had a Master's degree''} $\to$ \texttt{\{"education": "Masters"\}}), the tool executes a prediction using the \textbf{tabular foundation-model teacher}, and the same model consumes the raw tool output to produce the final natural-language answer (requiring up to three cloud LLM calls per query).
\item \textbf{Hybrid architecture (\texttt{HybridAgent}).} The cloud model is restricted to a single call: tool selection plus structured scenario parsing via OpenAI function-calling schemas (field names and types are declared in the schema, constraining what the model can emit). The selected tool executes a prediction using the \textbf{distilled student} MLP. A local \textbf{Qwen2.5-3B-Instruct} SLM then formats the raw tool JSON and the user's question into the final answer, entirely on-premises.
\end{itemize}

The two pipelines are held identical in tool schema, system-prompt intent, and evaluation query set; the only manipulated variables are (a) which model performs the tabular prediction (teacher vs. distilled student) and (b) which model performs response generation (cloud LLM multi-turn vs. local SLM single-pass). This isolates the cost/latency/quality effect of the distillation-plus-hybrid-split design from confounds like prompt engineering differences.

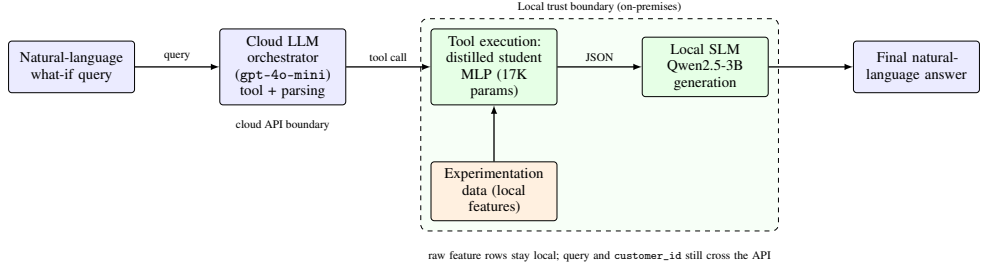
\begin{figure}[tbp]
\centering
\resizebox{0.92\textwidth}{!}{%
\begin{tikzpicture}[
  box/.style={draw, rounded corners=2pt, align=center, minimum height=1.1cm, inner sep=4pt, text width=2.4cm, font=\footnotesize},
  cloudc/.style={box, fill=blue!8},
  localc/.style={box, fill=green!10},
  datac/.style={box, fill=orange!12},
  flow/.style={-latex, thick},
  fnote/.style={font=\scriptsize, align=center}
]
\node[cloudc] (user) {Natural-language\\ what-if query};
\node[cloudc, right=1.8cm of user] (llm) {Cloud LLM\\ orchestrator\\ (\texttt{gpt-4o-mini})\\ tool + parsing};
\node[localc, right=1.8cm of llm] (tool) {Tool execution:\\ distilled student\\ MLP (17K params)};
\node[localc, right=1.8cm of tool] (slm) {Local SLM\\ Qwen2.5-3B\\ generation};
\node[cloudc, right=1.8cm of slm] (answer) {Final natural-\\ language answer};
\node[datac, below=1.2cm of tool] (data) {Experimentation\\ data (local features)};
\begin{scope}[on background layer]
\node[fit=(tool)(data)(slm), draw, dashed, rounded corners, inner sep=6pt, fill=green!3,
      label={[font=\scriptsize]above:Local trust boundary (on-premises)}] (bound) {};
\end{scope}
\draw[flow] (user) -- (llm) node[midway, fnote, above] {query};
\draw[flow] (llm) -- (tool) node[midway, fnote, above] {tool call};
\draw[flow] (tool) -- (slm) node[midway, fnote, above] {JSON};
\draw[flow] (data) -- (tool);
\draw[flow] (slm) -- (answer);
\node[fnote, below=0.2cm of llm] {cloud API boundary};
\node[fnote, below=0.3cm of bound, text width=7.5cm] {raw feature rows stay local; query and \texttt{customer\_id} still cross the API};
\end{tikzpicture}%
}
\caption{Hybrid architecture: a single cloud-LLM call performs tool selection and schema-constrained scenario parsing; the distilled student MLP runs the tabular prediction and a local Qwen2.5-3B-Instruct SLM generates the final answer on-premises. Raw 14-column feature rows do not reach third-party APIs; free-text queries and \texttt{customer\_id} still do.}
\label{fig:system}
\end{figure}

Figure~\ref{fig:system} illustrates the hybrid architecture's data flow: the cloud LLM processes only the user's free-text query and tool schemas, while the distilled student and local SLM, along with the underlying customer records, reside within the local trust boundary.

\paragraph{Motivation for the Asymmetric Split.} Section~\ref{sec:slm-ablation} shows that using the local SLM (Qwen2.5-3B-Instruct) zero-shot for scenario-to-JSON extraction caps tool-call accuracy at 58\%, compared to 98\% when using schema-constrained function calling with the cloud model (Section~\ref{sec:hybrid-results}). The architecture therefore maintains an asymmetric division of labor: the cloud LLM handles structured decision-making and extraction where small open models remain unreliable, whereas numeric prediction and text formatting are hosted on-premises.

\subsection{Observability}
\label{sec:observability}

A production deployment requires auditing \emph{why} a given decision was reached without re-running offline evaluations. We instrument both pipelines with LangSmith \citep{langsmith2024tracing} tracing (\codepath{scripts/observability.py}) as an implementation layer, using a lightweight decorator that adds zero overhead when disabled. Root spans capture orchestrator LLM token usage, prediction latencies (for both foundation teacher and student MLP), and local SLM generation. Offline evaluation metrics (tool-call accuracy, parsing recall, LLM-judge triad scores) are written back onto each live trace via \mbox{\texttt{Client.create\_feedback}}, enabling bidirectional debugging: operators can filter production runs by evaluation scores, and auditors can inspect end-to-end trace spans for any query. Tracing payload schemas and governance implications (span-level redaction for on-premises compliance) are detailed in Appendix~\ref{sec:appendix-prompts-params}.

\section{Distillation Benchmarking on Structured Data}
\label{sec:distillation-results}

\subsection{Datasets}

\textbf{Primary experimentation data.} \codepath{data/loan_data_with_max_loan.csv} is built by \codepath{scripts/prepare_business_data.py} from the UCI Adult census dataset \citep{kohavi1996adult} (48{,}842 rows, 23.9\% positive rate). The binary decision label is the real Adult income label (\texttt{>50K} $\to$ 1), so classification metrics are directly comparable to published Adult-benchmark numbers. \texttt{max\_loan} is a synthetic, business-rule-simulated regression target (multiplicative factors on age, workclass, education, occupation, net capital gain/loss, and hours worked) assigned only to rows with a positive outcome (11{,}687 rows, range \$90K-\$300K); it has no external ground truth and serves as a synthetic experimental benchmark target. Training used a stratified 10{,}000-row subsample (the tabular foundation-model teacher's pretrained context window is optimized for $\leq$10K rows) with an 80/20 train/test split (8{,}000/2{,}000).

\textbf{OpenML benchmark datasets.} To test generalization beyond the single business simulation, we ran the identical distillation protocol (\codepath{scripts/run_classification_benchmark.py}, reusing the same student architecture, loss, and seed as above) across five diverse OpenML classification benchmarks: three from the OpenML-CC18 \citep{bischl2021openml} suite (\textbf{credit-g}, German Credit, 1{,}000 rows, 800/200 split \citep{hofmann1994germancredit}; \textbf{bank-marketing}, Bank Marketing, 45{,}211 rows, stratified-subsampled to 10{,}000, 8{,}000/2{,}000 split \citep{moro2014bankmarketing}; and \textbf{kr-vs-kp}, Chess King-Rook vs.\ King-Pawn, 3{,}196 rows, 2{,}556/640 split \citep{shapiro1987krvskp}) alongside two high-dimensional or imbalanced sets (\textbf{internet-ads}, 3{,}279 rows, 1{,}558 features \citep{kushmerick1999internetads}; and \textbf{speeddating}, 8{,}378 rows, 118 features \citep{speeddating2006}). Full dataset characteristics are detailed in Table~\ref{tab:sources} (Appendix~\ref{sec:appendix-extended-classification}).

\subsection{Training protocol}
\label{sec:protocol}

Unless a caption names an exception, every classification result in Tables~\ref{tab:results} and~\ref{tab:extended-metrics} uses the defaults of \codepath{scripts/run_distillation.py} and \codepath{scripts/run_classification_benchmark.py}. There is no hint or embedding loss in those scripts.

\begin{table}[tbp]
\centering
\footnotesize
\begin{tabular}{ll}
\toprule
\textbf{Setting} & \textbf{Value} \\
\midrule
Classification $\alpha$ / $T$ / epochs & $0.7$ / $2.0$ / $10$ \\
Hint-loss weight & none (not implemented) \\
Optimizer / batch / seed & Adam, $10^{-3}$; batch 64; seed 42 \\
Primary regression (Section~\ref{sec:m1-regression}) & $\alpha=0.7$, $T=1.0$, 10 epochs \\
OpenML regression (Appendix~\ref{sec:appendix-extended-regression}) & $\alpha=0.9$, $T=1.0$, 200 epochs \\
$\alpha$-sweep (Table~\ref{tab:m1control}) & $T=3.0$ (kr-vs-kp: $T=2.0$, 5-fold CV) \\
\bottomrule
\end{tabular}
\caption{Canonical training configuration used for the reported tables. The $\alpha$-sweep in Table~\ref{tab:m1control} varies only the mixing weight around this protocol.}
\label{tab:protocol}
\end{table}

\subsection{Teacher and student architectures}

The teacher is \texttt{TabPFNClassifier} / \texttt{TabPFNRegressor} (\texttt{tabpfn==8.2.0}, in-context learning, no gradient fitting on the target dataset beyond context conditioning). The student is a compact feed-forward network, identical in shape across every dataset and task, with only the input width varying by the one-hot-encoded feature count:
\begin{equation}
g_\phi(x) = W_3\,\sigma_2\big(\operatorname{Dropout}_{0.2}(W_2\,\sigma_1(\operatorname{Dropout}_{0.2}(W_1 x + b_1)) + b_2)\big) + b_3
\end{equation}
with $W_1 \in \mathbb{R}^{64\times d_{\text{in}}}$, $W_2\in\mathbb{R}^{32\times64}$, $W_3\in\mathbb{R}^{C\times32}$ ($\sigma_1=\mathrm{ReLU}$; $\sigma_2=\mathrm{sigmoid}$ for the classifier head, $\mathrm{ReLU}$ for the regressor head; $C=2$ for classification, $C=1$ for regression). Both teacher and student are trained/evaluated with seed 42 against teacher-produced soft logits precomputed once per training run (remaining hyperparameters in Table~\ref{tab:protocol}). A layer-by-layer accounting of all student MLP weights is detailed in Appendix~\ref{sec:appendix-prompts-params}.

\subsection{Parameter-count accounting}
\label{sec:param-table}

Parameter counts were measured directly from the loaded PyTorch modules (teacher: \texttt{sum(p.numel() for p in model.parameters())} on the fitted \texttt{tabpfn} model; student: from saved checkpoints). Table~\ref{tab:params} (Appendix~\ref{sec:appendix-extended-classification}) reports full parameter counts and compression ratios; Figure~\ref{fig:params} visualizes them on a log scale. On the primary loan task, the combined predictor achieves a \textbf{6,532$\times$} parameter reduction (111,428,088 teacher parameters to 17,059 student parameters: 8,546 for classification, 8,513 for regression). Across OpenML-CC18 classification benchmarks, compression reaches \textbf{10,063$\times$} (credit-g, 5,282 parameters), \textbf{10,852$\times$} (bank-marketing, 4,898 parameters), and \textbf{11,611$\times$} (kr-vs-kp, 4,578 parameters). Smaller-input datasets exhibit higher compression ratios because the student's first layer scales linearly with one-hot feature dimension ($64d_{\text{in}}+2{,}210$) while the teacher's stored parameter count remains fixed.

\subsection{Accuracy and AUC retention}

Table~\ref{tab:results} reports teacher and student test-set metrics (visualized in Figure~\ref{fig:acc-auc}, Appendix~\ref{sec:appendix-extended-classification}). \emph{Retention} is defined as $\text{student metric} / \text{teacher metric}$. Because some held-out test sets are small, sampling noise can make the student's measured metric nominally exceed the teacher's; we therefore report retention uncapped. Across all six datasets, accuracy retention ranges 95.4--100.5\% and AUC retention ranges 96.8--100.0\%, at a $6{,}220\times$ (business-simulation classifier) to $11{,}611\times$ (kr-vs-kp) parameter reduction (Table~\ref{tab:params}). Extended classification sweeps covering precision, recall, and F1 on the same checkpoints are in Appendix~\ref{sec:appendix-extended-classification}. These retention figures are evaluated against hard-label baselines in Section~\ref{sec:m1-control} to isolate the contribution of teacher soft targets.

\begin{table}[tbp]
\centering
\footnotesize
\resizebox{\textwidth}{!}{%
\begin{tabular}{lrrrrrr}
\toprule
\textbf{Dataset} & \textbf{Teacher acc.} & \textbf{Student acc.} & \textbf{Acc. retention} & \textbf{Teacher AUC} & \textbf{Student AUC} & \textbf{AUC retention} \\
\midrule
Loan approval & 0.8495 & 0.8450 & 99.5\% & 0.9175 & 0.9125 & 99.5\% \\
credit-g & 0.7650 & 0.7300 & 95.4\% & 0.7857 & 0.7714 & 98.2\% \\
bank-marketing & 0.9070 & 0.8965 & 98.8\% & 0.9263 & 0.8970 & 96.8\% \\
internet-ads & 0.9604 & 0.9649 & 100.5\% & 0.9858 & 0.9845 & 99.9\% \\
speeddating & 0.8586 & 0.8598 & 100.1\% & 0.8653 & 0.8653 & 100.0\% \\
kr-vs-kp & 0.9969 & 0.9672 & 97.0\% & 0.9999 & 0.9948 & 99.5\% \\
\bottomrule
\end{tabular}%
}
\caption{Teacher vs.\ student classification quality (same checkpoints as Table~\ref{tab:extended-metrics}; protocol in Table~\ref{tab:protocol}: $\alpha=0.7$, $T=2.0$, 10 epochs, no hint-loss, seed 42). Retention is student/teacher. Precision, recall, and F1 for these runs are in Table~\ref{tab:extended-metrics}.}
\label{tab:results}
\end{table}

\subsection{Hard-Label-Only Control Ablation}
\label{sec:m1-control}

To isolate the supervisory contribution of teacher soft targets from the baseline inductive capacity of the MLP architecture, we evaluate an identical student architecture trained solely on ground-truth hard labels ($\alpha=0$). Hyperparameters, training epochs, and random seeds are held fixed to Table~\ref{tab:protocol}. We evaluate test ROC-AUC across all six datasets (Table~\ref{tab:m1control}, Appendix~\ref{sec:appendix-extended-classification}).

Across all six datasets, the hard-label baseline ($\alpha=0$) achieves the lowest performance, with ROC-AUC increasing as $\alpha$ increases toward $0.7$. At the canonical $\alpha=0.7$, teacher soft targets yield gains over hard labels alone of 0.0210 (loan approval), 0.0589 (credit-g), 0.0702 (bank-marketing), 0.0423 (internet-ads), 0.0254 (speeddating), and 0.0296 (kr-vs-kp, 5-fold CV), i.e.\ 2.1--7.0 AUC points. This confirms that teacher soft targets transmit inductive bias beyond what is captured by ground-truth labels.

\subsection{Regression Head Evaluation and Soft-Target Control}
\label{sec:m1-regression}

For the regression head on the primary experimentation data, the teacher achieves RMSE \$75{,}737 (WMAPE 0.712) on the test set. Evaluated conditionally on its own approval decisions (the deployed operational setting; \codepath{scripts/prediction_utils.py}), the student achieves RMSE \$88{,}115 and WMAPE 0.873 (424 predicted approvals out of 2{,}000 test rows; unconditional evaluation yields RMSE \$84{,}717, WMAPE 1.213). Evaluating the teacher regressor on the student's approved subset yields WMAPE 0.570 and RMSE \$105{,}422. We report both metrics to reflect typical-loan error (WMAPE) alongside sensitivity to large-balance deviations (RMSE).

Sweeping $\alpha \in \{0, 0.3, 0.7, 1.0\}$ (Eq.~\ref{eq:kd-reg}, $T=1.0$) on the unconditional test set yields WMAPE values of $1.266 \to 1.226 \to 1.160 \to 1.107$, showing that pure soft-target regression ($\alpha=1.0$) achieves the lowest error, consistent with the classification findings (see Figure~\ref{fig:loss} in Appendix~\ref{sec:appendix-extended-classification}). Together with Section~\ref{sec:m1-control}, this confirms that teacher supervision benefits both task heads. Extended evaluations across five additional OpenML regression benchmarks are presented in Appendix~\ref{sec:appendix-extended-regression}.

\section{Hybrid Architecture: Token Cost, Latency, and Quality}
\label{sec:hybrid-results}

We evaluate the full-LLM and hybrid pipelines of Section~\ref{sec:system} on the same 50-query natural-language what-if benchmark (\codepath{data/hybrid_scenario_eval_50.jsonl}; 30 of the 50 queries carry an explicit scenario-modification ground truth) on three hardware profiles (full list in Appendix~\ref{sec:appendix-reproducibility}): (a)~a shared 32GB x86\_64 CPU host (PyTorch CPU); (b)~the same class of shared host with an NVIDIA RTX 4060 Ti 16GB (\texttt{bfloat16} for local Qwen; TabPFN also on GPU); (c)~a dedicated Apple M5 Pro (48GB, CPU backend only; Table~\ref{tab:costlatm5}). Profiles (a) and (c) are both CPU execution, not the same machine: (a) is a contended development host and (c) is a clean-room check. All three use the same cloud endpoint (\texttt{gpt-4o-mini}) for orchestration, so \emph{token cost} and tool-call/judge accuracy are hardware-invariant up to run-to-run token noise. Wall-clock latency is not: local TabPFN, local Qwen, and host/network contention change with the box. The student forward pass remains sub-millisecond even on CPU.

\subsection{Tool-call and parsing accuracy}

\begin{table}[tbp]
\centering
\begin{minipage}{0.54\textwidth}
\centering
\scriptsize
\begin{tabular}{lrrr}
\toprule
\textbf{Metric} & \textbf{Hybrid} & \textbf{Full-LLM} & \textbf{Diff.} \\
\midrule
Tool-call accuracy & 0.98 & 0.90 & +0.08 \\
Parameter accuracy & 0.98 & 0.90 & +0.08 \\
Scenario parsing recall & 0.894 & 0.717 & +0.177 \\
Scenario parsing precision & 0.906 & 0.717 & +0.189 \\
Errored queries (of 50) & 0 & 0 & -- \\
\bottomrule
\end{tabular}
\caption{Structured extraction quality (50 queries).}
\label{tab:toolcall}
\end{minipage}\hfill
\begin{minipage}{0.43\textwidth}
\centering
\scriptsize
\begin{tabular}{lrr}
\toprule
\textbf{Metric (mean)} & \textbf{Hybrid} & \textbf{Full-LLM} \\
\midrule
Faithfulness & 0.750 & 0.830 \\
Context relevance & 0.780 & 1.000 \\
Answer relevance & 0.940 & 1.000 \\
\midrule
\multicolumn{3}{l}{\textit{Zero errors on either pipeline.}} \\
\\
\bottomrule
\end{tabular}
\caption{LLM-as-judge answer scores.}
\label{tab:judge}
\end{minipage}
\end{table}

\textbf{Reliability accounting.} Both pipelines answer all 50 benchmark queries without error (defined as unhandled exceptions post-selection returning no response); evaluations in this section are computed over the full 50-query set for both pipelines. Tool-call accuracy, parsing quality, and LLM-judge scores are hardware-independent.

Table~\ref{tab:toolcall} and Table~\ref{tab:slmablation} are not interchangeable. Table~\ref{tab:toolcall} scores the end-to-end hybrid and full-LLM agents on the 50-query benchmark when the cloud model uses OpenAI function calling with named parameters (\texttt{customer\_id}, \texttt{modifications}, and declared field types) bound in the tool schema; that is the production interface, and it is why hybrid tool-call accuracy reaches 0.98 versus 0.90 for unconstrained multi-turn extraction in the full-LLM baseline. Table~\ref{tab:slmablation} is a standalone parser ablation (\codepath{scripts/run_parse_accuracy_eval.py}) that gives \emph{both} \texttt{gpt-4o-mini} and Qwen2.5-3B the same generic JSON schema without those named declarations, decoded locally via Outlines. Under that weaker interface the cloud model drops to 0.38 exact-match accuracy and the local SLM reaches only 0.58---which is why the deployed hybrid architecture keeps schema-constrained cloud function calling for parsing rather than replacing it with the 3B SLM.

\subsection{LLM-judge answer quality}

Final answers are scored by an LLM judge (\texttt{gpt-4o-mini}, structured output) on the RAGAS-style faithfulness / context-relevance / answer-relevance triad \citep{es2024ragastriad} against the tool output consumed and the user's question (Table~\ref{tab:judge}). Faithfulness favors the full-LLM pipeline (0.83 vs.\ 0.75); context- and answer-relevance are lower for the hybrid pipeline, because the local Qwen2.5-3B formatter occasionally surfaces raw feature keys (e.g., \texttt{"educationserie"}, malformed unicode in \texttt{occupation}) without fully synthesizing the user's specific query, a known limitation of small, generic language models without task-specific fine-tuning.

\subsection{Cost and latency}

Total cost across all 50 queries is \$0.00479 for the hybrid pipeline versus \$0.01011 for the full-LLM pipeline, representing a \textbf{2.11$\times$} reduction. This difference is driven by cloud LLM call frequency (a single constrained call in the hybrid architecture versus up to three unconstrained calls in the baseline), not by CPU vs.\ GPU. Table~\ref{tab:costlat} therefore repeats the same per-query dollars under the GPU headers. The M5 Pro run (Table~\ref{tab:costlatm5}) is an independent 50-query pass, so its pair (\$0.00475 vs.\ \$0.01021, 2.15$\times$) differs only by token-count noise.

\textbf{Hardware amortization and total cost.} This $2.11\times$ saving reflects marginal cloud API token spend (\$0.106 saved per 1{,}000 queries). Amortizing a dedicated \$500 GPU (RTX 4060 Ti) requires $\approx$4.71M queries before breaking even against cloud savings alone (excluding power and maintenance). Thus, marginal cost savings apply at enterprise scale or on existing infrastructure, while latency and governance benefits apply unconditionally from query one.

\begin{table}[tbp]
\centering
\footnotesize
\resizebox{\textwidth}{!}{%
\begin{tabular}{lrrrr}
\toprule
 & \textbf{Hybrid (CPU)} & \textbf{Full-LLM (CPU)} & \textbf{Hybrid (GPU)} & \textbf{Full-LLM (GPU)} \\
\midrule
Mean cost / query & \$9.59$\times10^{-5}$ & \$2.02$\times10^{-4}$ & \$9.59$\times10^{-5}$ & \$2.02$\times10^{-4}$ \\
Mean orchestrator latency\textsuperscript{\dag} & 12.75s & 5.91s & 0.85s & 0.39s \\
Mean formatter/gen. latency & 10.15s & 1.57s & 0.68s & 0.10s \\
Mean ML predict latency & 5.67ms & 12{,}578ms & 0.38ms & 838.6ms \\
Unaccounted (tool-exec.\ wall time)\textsuperscript{*} & $\approx$0.00s & 67.88s & $\approx$0.00s & 4.53s \\
\midrule
Mean total latency (measured) & 22.91s & 87.95s & 1.53s & 5.86s \\
\bottomrule
\end{tabular}%
}
\caption{Cost and latency, 50-query benchmark on two \emph{shared} development hosts: 32GB x86\_64 CPU vs.\ RTX 4060 Ti 16GB. Cloud token cost is identical across these columns (same \texttt{gpt-4o-mini} run). ``ML predict latency'' is the tabular-model call only (student MLP for hybrid, TabPFN teacher for full-LLM). Mean total latency is measured wall-clock per query. \textsuperscript{\dag}Orchestrator is the same cloud API on every profile; 12.75s vs.\ 0.85s is wait time on a loaded CPU host (queueing/network), not a different model. \textsuperscript{*}For the full-LLM pipeline, the three instrumented spans do not exhaust the wall-clock window; the residual tracks host contention during teacher in-context inference. A dedicated third profile (Apple M5 Pro) is Table~\ref{tab:costlatm5}.}
\label{tab:costlat}
\end{table}

The hybrid pipeline is \textbf{3.84$\times$} faster end-to-end on the shared 32GB CPU host (22.91s vs. 87.95s), \textbf{3.83$\times$} faster on the RTX 4060 Ti (1.53s vs. 5.86s), and \textbf{4.82$\times$} faster on the dedicated Apple M5 Pro CPU (4.96s vs. 23.94s, Table~\ref{tab:costlatm5}). Absolute seconds are not comparable across the two CPU columns: the M5 Pro is a different, unloaded machine, not a faster run of the 32GB host.

\textbf{Latency attribution (existing spans; no new runs).} The hybrid vs.\ full-LLM comparison changes two factors at once: the tabular backend (TabPFN teacher vs.\ student MLP) and the generation path (up to three cloud LLM calls vs.\ one schema-constrained cloud call plus local Qwen). Table~\ref{tab:latency-attr} regroups the already-instrumented rows of Tables~\ref{tab:costlat} and~\ref{tab:costlatm5}. On every hardware profile, TabPFN inference plus the unaccounted residual---which we attribute to host contention during in-context teacher calls---accounts for 90--92\% of full-LLM wall-clock time (80.46s of 87.95s on CPU; 5.37s of 5.86s on GPU; 21.55s of 23.94s on M5 Pro). Replacing that backend with the student drops the same bucket to milliseconds. By contrast, orchestrator plus formatter time is \emph{higher} in the hybrid pipeline (22.90s vs.\ 7.48s on CPU; 1.53s vs.\ 0.49s on GPU; 4.96s vs.\ 2.40s on M5 Pro), because local Qwen formatting is slower than the baseline's cloud generation even though cloud round-trips are fewer. The 2.1$\times$ token-cost reduction is therefore the right quantity to attribute to the fewer-LLM-call split; the 3.8--4.8$\times$ latency reduction is not. A crossed 2$\times$2 (student with cloud formatter; teacher with local formatter) was not run; the decomposition above is additive over measured spans, not a factorial experiment.

\begin{table}[tbp]
\centering
\footnotesize
\begin{tabular}{lrrrrrr}
\toprule
 & \multicolumn{2}{c}{\textbf{Shared CPU}} & \multicolumn{2}{c}{\textbf{Shared GPU}} & \multicolumn{2}{c}{\textbf{Dedicated M5}} \\
\cmidrule(lr){2-3}\cmidrule(lr){4-5}\cmidrule(lr){6-7}
\textbf{Mean seconds / query} & Hybrid & Full-LLM & Hybrid & Full-LLM & Hybrid & Full-LLM \\
\midrule
Tabular predict + residual & 0.01 & 80.46 & 0.00 & 5.37 & 0.02 & 21.55 \\
\quad share of full-LLM total & -- & 91.5\% & -- & 91.6\% & -- & 90.0\% \\
Orchestrator + formatter & 22.90 & 7.48 & 1.53 & 0.49 & 4.96 & 2.40 \\
\midrule
Measured total & 22.91 & 87.95 & 1.53 & 5.86 & 4.96 & 23.94 \\
\bottomrule
\end{tabular}
\caption{Additive latency attribution from instrumented spans in Tables~\ref{tab:costlat} and~\ref{tab:costlatm5} (no additional evaluation runs). ``Tabular predict + residual'' is student MLP + $\approx$0 in the hybrid columns and TabPFN + unaccounted wall time in the full-LLM columns. Hybrid orchestrator+formatter exceeds the full-LLM LLM path on all three profiles.}
\label{tab:latency-attr}
\end{table}

The largest instrumented single-call gap remains the tabular prediction: the distilled student (17{,}059 parameters total) answers in 5.67ms (CPU host), 0.38ms (GPU), and 6.09ms (M5 Pro), while the teacher takes 12{,}578ms, 838.6ms, and 19{,}579ms (2{,}219$\times$, 2{,}207$\times$, and 3{,}213$\times$; Figure~\ref{fig:predict}, Appendix~\ref{sec:appendix-reproducibility}). On dedicated hardware (Apple M5 Pro), the full-LLM residual shrinks from 67.88s to 1.97s (8\% of total), corroborating the contention hypothesis. Real wall-clock latency that full-LLM deployments pay on shared infrastructure reinforces the necessity of removing in-context foundation models from the per-query hot path.

\section{Discussion: Production Hardening and Data Governance}
\label{sec:discussion}

\textbf{Distillation Versus Architectural Downscaling.} The latency of tabular foundation models stems fundamentally from in-context attention over exemplars. Reducing exemplar volume directly degrades inductive accuracy on the target distribution. In contrast, distilling into a feed-forward student eliminates attention overhead during serving, providing a $\ge$2,200$\times$ per-query inference speedup following an offline training step of under two minutes. As established in Section~\ref{sec:distillation-results}, the student matches teacher accuracy while improving upon hard-label training by 2.1--7.0 AUC points (Table~\ref{tab:m1control}), demonstrating inductive transfer from teacher soft probabilities.

\textbf{Response Formatting Fidelity.} Context- and answer-relevance scores are lower for the hybrid pipeline (Table~\ref{tab:judge}: 0.78/0.94 vs.\ 1.00/1.00) because the local Qwen2.5-3B formatter is not task-fine-tuned, occasionally surfacing raw feature keys in final answers. Constraining the formatter's output schema or applying parameter-efficient fine-tuning can address this disparity. Both pipelines achieved 100\% execution reliability across all 50 benchmark queries (Table~\ref{tab:toolcall}).

\textbf{Subgroup Fairness Under Compression.} Soft-target compression does not guarantee preservation of group-conditional error rates even when aggregate metrics are maintained \citep{hooker2020characterising}. Auditing demographic parity and equalized-odds differences on the Adult test set reveals that while sex disparities remain stable (demographic parity 0.180$\to$0.176; equalized odds 0.093$\to$0.080), race disparities widen under distillation: demographic parity increases 0.240$\to$0.256 and equalized-odds difference widens 0.431$\to$0.490 (+14\% relative). Although true-positive rates improve across individual racial subgroups under the student (e.g., Black applicants: 0.474$\to$0.632), non-uniform improvement rates widen the disparate-impact gap. Because aggregate retention metrics can mask disparate subgroup effects, fairness audits are essential when deploying compressed models in regulated domains (detailed protocol in Appendix~\ref{sec:appendix-validity}).

\textbf{Data Governance.} In the full-LLM baseline, intermediate tool outputs (including demographic and financial attributes) enter the cloud LLM prompt context across multiple turns. In the hybrid architecture, raw 14-column feature rows and model predictions remain inside the local trust boundary. This is data minimization, not elimination of personal data from the cloud path: the orchestrator still receives the user's free-text query and the integer \texttt{customer\_id} required to select a tool. Both are personal data under GDPR Art.~4(1) \citep{regulation2016gdpr,voigt2017gdpr} if they identify or are linkable to a natural person. Localizing the feature row therefore reduces, but does not remove, Art.~5(1)(c) exposure. Extended compliance accounting is detailed in Appendix~\ref{sec:appendix-validity}.

\textbf{Observability and Auditability.} Regulated decision-making requires operational auditability alongside offline benchmark evaluations. As established in Section~\ref{sec:observability}, attaching evaluative scores directly to LangSmith trace trees provides bidirectional verification: compliance auditors can inspect intermediate tool parameters, student/teacher prediction differences, and formatter generation directly from production spans (Appendix~\ref{sec:appendix-prompts-params}).

\subsection{Limitations}
\label{sec:limitations}
The following constraints bound how far the reported numbers should be generalized; Appendix~\ref{sec:appendix-validity} expands each item.
\begin{enumerate}
\item \textbf{Orchestrator tier.} \texttt{gpt-4o-mini} is a proxy for a cheaper \texttt{gpt-4o-nano} production tier. Schema adherence should transfer; absolute token prices and orchestrator latency will not.
\item \textbf{What-if sample size.} The 50-query benchmark separates pipelines on latency and tool-call accuracy, but it is underpowered for subtle natural-language or LLM-judge differences.
\item \textbf{Synthetic regression label.} \texttt{max\_loan} is a rule-simulated target with no external ground truth; classification labels are the real Adult income indicator.
\item \textbf{Hardware and tracing residual.} The 32GB CPU and RTX 4060 Ti sweeps were collected on shared development machines; the M5 Pro is a dedicated reproduction. Orchestrator latency of 12.75s (shared CPU) vs.\ $\approx$0.85s (GPU and M5) is the same \texttt{gpt-4o-mini} endpoint under host/network contention, not a different model. The large unaccounted full-LLM residual (67.88s CPU) shrinks to 1.97s on M5 Pro, consistent with contention during TabPFN in-context inference. Latency attribution in Table~\ref{tab:latency-attr} uses these spans additively; a factorial 2$\times$2 (student with cloud formatter; teacher with local formatter) was not run.
\item \textbf{Single seed.} Distillation tables use seed 42. Multi-seed weight-initialization sweeps are future work.
\item \textbf{Fairness scope.} Subgroup audits cover sex and race on Adult groups with $\ge 20$ test rows only.
\end{enumerate}

\section{Conclusion}

We distilled a TabPFN teacher into a feed-forward student, retaining 95.4--100.5\% accuracy and 96.8--100.0\% AUC on the Table~\ref{tab:results} checkpoints (6{,}220$\times$ for the 8{,}546-parameter classifier; 6{,}532$\times$ for the 17{,}059-parameter two-head loan pipeline). Hard-label controls ($\alpha=0$) confirm that teacher soft targets yield a 2.1--7.0 AUC point gain alongside monotonic regression improvements, demonstrating that the teacher provides supervisory signal beyond architecture capacity. Embedding the distilled student in a hybrid LangGraph architecture (restricting cloud LLM calls to tool selection while serving tabular prediction and response formatting locally) reduces end-to-end latency by 3.8--4.8$\times$, almost entirely by removing TabPFN from the hot path; fewer cloud calls cut token cost 2.1$\times$ rather than wall-clock time. Tool-call accuracy improves (0.98 vs.\ 0.90). Raw feature rows stay on-premises while free-text queries and \texttt{customer\_id} still leave the local boundary. Hardware amortization of a dedicated GPU still requires multi-million-query scale. Standalone ablations demonstrate that local SLMs achieve 58\% parsing accuracy zero-shot under a generic schema, supporting the asymmetric division of labor. Subgroup fairness audits reveal that while sex disparities remain stable, equalized-odds differences by race widen by 14\% under compression despite individual TPR gains, highlighting the necessity of subgroup evaluations in model distillation workflows. Future work includes fine-tuning the local formatter, scaling to unlabelled transfer sets, and deploying span-level redaction for hybrid observability.

\bibliographystyle{plainnat}
\bibliography{references}

\newpage
\appendix

\section{Extended Classification Benchmark Sweeps}
\label{sec:appendix-extended-classification}
\label{sec:extended-metrics}

To broaden coverage beyond the datasets reported in Table~\ref{tab:results} and probe class-imbalanced and harder settings, we report precision, recall, and F1 for the \emph{same} 1-fold checkpoints (\codepath{scripts/run_classification_benchmark.py}, Table~\ref{tab:protocol}: seed 42, 80/20 split, $\alpha=0.7$, $T=2.0$, 10 epochs, no hint-loss). Table~\ref{tab:sources} reports each dataset's origin and raw dimensions; Table~\ref{tab:extended-metrics} reports the full metric set.

\begin{table}[tbp]
\centering
\footnotesize
\resizebox{\textwidth}{!}{%
\begin{tabular}{lllrrr}
\toprule
\textbf{Dataset} & \textbf{Source} & \textbf{OpenML id} & \textbf{Instances} & \textbf{Raw features} & \textbf{Positive rate} \\
\midrule
Loan approval & UCI Adult census \citep{kohavi1996adult} & 1590 & 48{,}842 (10{,}000 used) & 14 & 23.9\% \\
credit-g & German Credit \citep{hofmann1994germancredit} & 31 & 1{,}000 & 20 & 30.0\% \\
bank-marketing & Bank Marketing \citep{moro2014bankmarketing} & 1461 & 45{,}211 (10{,}000 used) & 16 & 11.7\% \\
internet-ads & Internet Advertisements \citep{kushmerick1999internetads} & 1554 & 3{,}279 & 1{,}558 & 14.0\% \\
speeddating & Speed Dating Experiment \citep{speeddating2006} & 40536 & 8{,}378 & 118\textsuperscript{*} & 16.5\% \\
kr-vs-kp\textsuperscript{\dag} & Chess (King-Rook vs.\ King-Pawn) \citep{shapiro1987krvskp} & 3 & 3{,}196 & 36 & 52.2\% \\
\bottomrule
\end{tabular}%
}
\caption{Data sources and dimensions for the extended comparison. ``Raw features'' counts columns before one-hot encoding. Datasets exceeding the teacher's $\leq$10K-row pretrained context window (loan, bank-marketing) are stratified-subsampled to 10{,}000 rows before the 80/20 split, consistent with Table~\ref{tab:results}. \textsuperscript{*}speeddating's raw OpenML export has 123 attributes; we removed \texttt{decision} and \texttt{decision\_o} (each participant's individual yes/no), whose logical AND reconstructs the \texttt{match} target with 100\% agreement, constituting a direct label leak, alongside a free-text \texttt{field} column (260 categories) and an \texttt{expected\_num\_interested\_in\_me} column that is 78.5\% missing; the remaining 118 columns' missing values (${\sim}$1.5\% of cells) were median/mode-imputed. \textsuperscript{\dag}kr-vs-kp is a sixth, harder-dataset check evaluated via \codepath{scripts/run_classification_benchmark.py} under its default benchmark configuration; its 36 raw features are all Boolean chess-board predicates (no missing values, no subsampling needed since it fits under the 10K-row context cap), and ``positive rate'' is the \texttt{won} (White-can-win) class share.}
\label{tab:sources}
\end{table}

\begin{table}[tbp]
\centering
\footnotesize
\resizebox{\textwidth}{!}{%
\begin{tabular}{lrrrrrrrrrr}
\toprule
& \multicolumn{5}{c}{\textbf{Teacher}} & \multicolumn{5}{c}{\textbf{Student}} \\
\cmidrule(lr){2-6} \cmidrule(lr){7-11}
\textbf{Dataset} & Acc. & AUC & Prec. & Rec. & F1 & Acc. & AUC & Prec. & Rec. & F1 \\
\midrule
Loan approval & 0.8495 & 0.9175 & 0.7181 & 0.6117 & 0.6607 & 0.8450 & 0.9125 & 0.7107 & 0.5950 & 0.6477 \\
credit-g & 0.7650 & 0.7857 & 0.8121 & 0.8643 & 0.8374 & 0.7300 & 0.7714 & 0.7756 & 0.8643 & 0.8176 \\
bank-marketing & 0.9070 & 0.9263 & 0.6263 & 0.5085 & 0.5613 & 0.8965 & 0.8970 & 0.5931 & 0.3675 & 0.4538 \\
internet-ads & 0.9604 & 0.9858 & 0.9559 & 1.0000 & 0.9775 & 0.9649 & 0.9845 & 0.9608 & 1.0000 & 0.9800 \\
speeddating & 0.8586 & 0.8653 & 0.6291 & 0.3442 & 0.4450 & 0.8598 & 0.8653 & 0.6614 & 0.3043 & 0.4169 \\
kr-vs-kp\textsuperscript{\dag} & 0.9969 & 0.9999 & 0.9970 & 0.9970 & 0.9970 & 0.9672 & 0.9948 & 0.9728 & 0.9641 & 0.9684 \\
\bottomrule
\end{tabular}%
}
\caption{Teacher vs.\ student classification quality on the same checkpoints as Table~\ref{tab:results} (protocol in Table~\ref{tab:protocol}). Precision/recall/F1 are reported for the positive class (loan approval, credit-g ``good'' credit, bank-marketing ``yes'' subscription, internet-ads ``ad'', speeddating ``match'', kr-vs-kp ``won''). On the two most imbalanced sets (bank-marketing 11.7\% positive, speeddating 16.5\% positive), both teacher and student show lower recall/F1 than accuracy/AUC indicates. The student is within 1--3 F1 points of the teacher on four of six datasets and behind on bank-marketing (F1 0.454 vs.\ 0.561).}
\label{tab:extended-metrics}
\end{table}

\begin{table}[tbp]
\centering
\small
\begin{tabular}{lrrrr}
\toprule
\textbf{Dataset} & \textbf{Input dim.} & \textbf{Teacher params} & \textbf{Student params} & \textbf{Compression} \\
\midrule
Loan approval (classifier) & 99 & 53{,}153{,}144 & 8{,}546 & 6{,}220$\times$ \\
Loan amount (regressor) & 99 & 58{,}274{,}944 & 8{,}513 & 6{,}844$\times$ \\
\textbf{Loan task, combined} & -- & \textbf{111{,}428{,}088} & \textbf{17{,}059} & \textbf{6{,}532$\times$} \\
\midrule
credit-g (classifier) & 48 & 53{,}153{,}144 & 5{,}282 & 10{,}063$\times$ \\
bank-marketing (classifier) & 42 & 53{,}153{,}144 & 4{,}898 & 10{,}852$\times$ \\
kr-vs-kp (classifier) & 37 & 53{,}153{,}144 & 4{,}578 & 11{,}611$\times$ \\
\bottomrule
\end{tabular}
\caption{Parameter counts (measured directly from loaded models) and compression ratios. The teacher's classifier architecture and parameter count are identical across datasets because the tabular foundation model performs in-context learning with fixed pretrained weights; only the input projection's effective receptive field over columns varies while the module's stored parameter count remains constant. ``Input dim.'' is the student's actual input width $d_{\text{in}}$ (where numeric columns pass through \texttt{StandardScaler} and categorical columns pass through \texttt{OneHotEncoder(drop=\textquotesingle first\textquotesingle)}, in contrast to raw column counts in Table~\ref{tab:sources}), since it is $d_{\text{in}}$ that sets the student's first-layer width $W_1\in\mathbb{R}^{64\times d_{\text{in}}}$ and hence its total parameter count ($64d_{\text{in}}+2{,}210$ for the two-class classifier head): e.g.\ credit-g's 20 raw columns one-hot-encode to 48 dimensions, while kr-vs-kp's 36 raw Boolean columns encode to 37. Visualized on log scale in Figure~\ref{fig:params}.}
\label{tab:params}
\end{table}

\begin{table}[tbp]
\centering
\footnotesize
\begin{tabular}{lrrrrrr}
\toprule
\textbf{Dataset} & \textbf{Teacher} & \textbf{$\alpha{=}0$} & \textbf{$\alpha{=}0.3$} & \textbf{$\alpha{=}0.7$} & \textbf{$\alpha{=}0.85$} & \textbf{$\alpha{=}1.0$} \\
\midrule
Loan approval & 0.9117 & 0.8883 & 0.9007 & 0.9093 & 0.9098 & 0.9103 \\
credit-g & 0.7857 & 0.7181 & 0.7765 & 0.7770 & 0.7777 & 0.7777 \\
bank-marketing & 0.9263 & 0.8274 & 0.8881 & 0.8976 & 0.8975 & 0.8974 \\
internet-ads & 0.9858 & 0.9434 & 0.9670 & 0.9857 & 0.9863 & 0.9853 \\
speeddating & 0.8653 & 0.8353 & 0.8562 & 0.8607 & 0.8624 & 0.8621 \\
kr-vs-kp\textsuperscript{\dag} & 0.9999 & 0.9652 & 0.9834 & 0.9948 & 0.9994 & 0.9989 \\
\bottomrule
\end{tabular}
\caption{Test AUC sweep over KD loss mixing weight $\alpha$ ($T=3.0$, seed 42; reported Tables~\ref{tab:results} and~\ref{tab:extended-metrics} use $\alpha=0.7$, $T=2.0$). $\alpha=0$ denotes the hard-label-only control (no teacher signal). \textsuperscript{\dag}kr-vs-kp reports mean test AUC from 5-fold cross-validation ($T=2.0$).}
\label{tab:m1control}
\end{table}

\begin{figure}[tbp]
\centering
\includegraphics[width=0.85\textwidth]{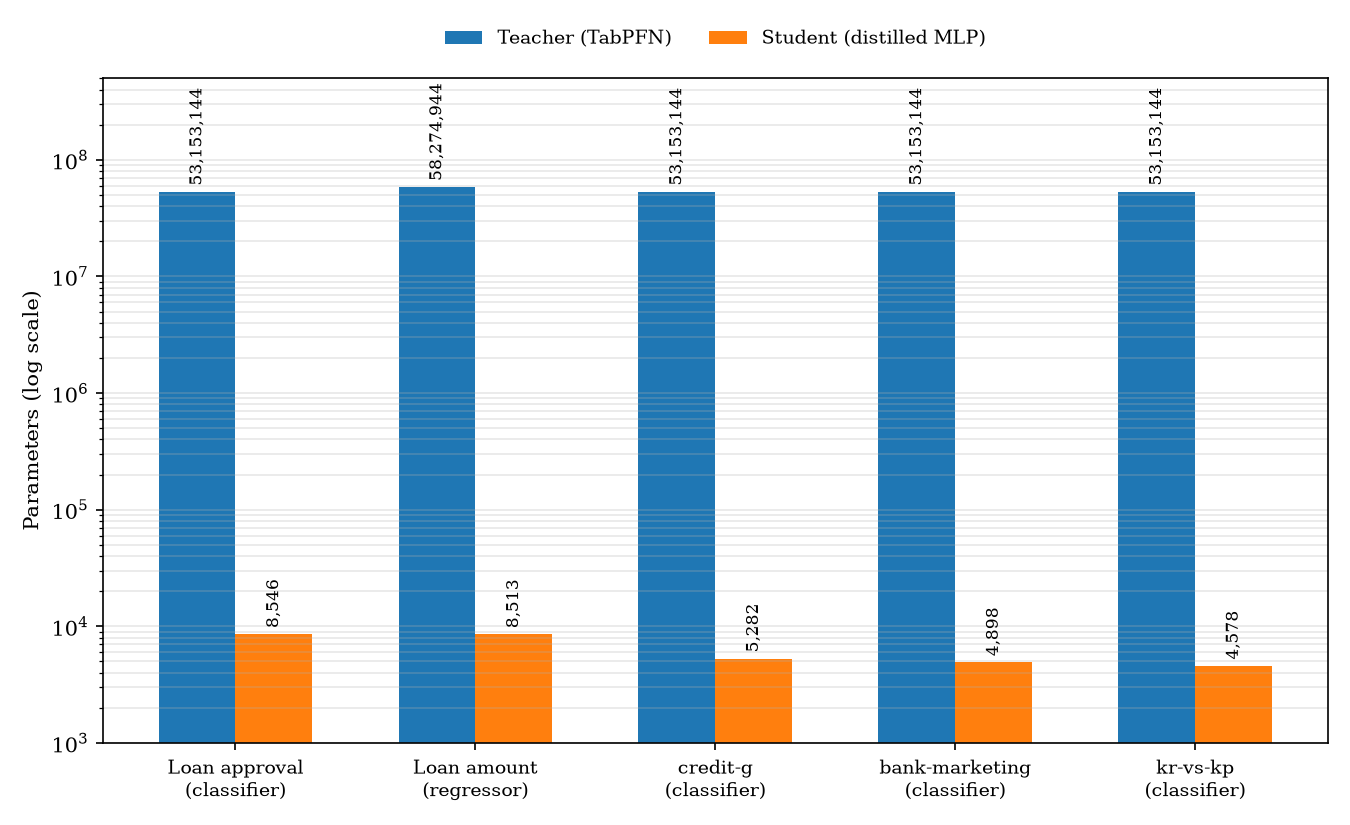}
\caption{Teacher (tabular foundation model) vs. student (distilled MLP) parameter count, log scale, across the primary business-decision simulation and three OpenML-CC18 datasets.}
\label{fig:params}
\end{figure}

\begin{figure}[tbp]
\centering
\includegraphics[width=0.95\textwidth]{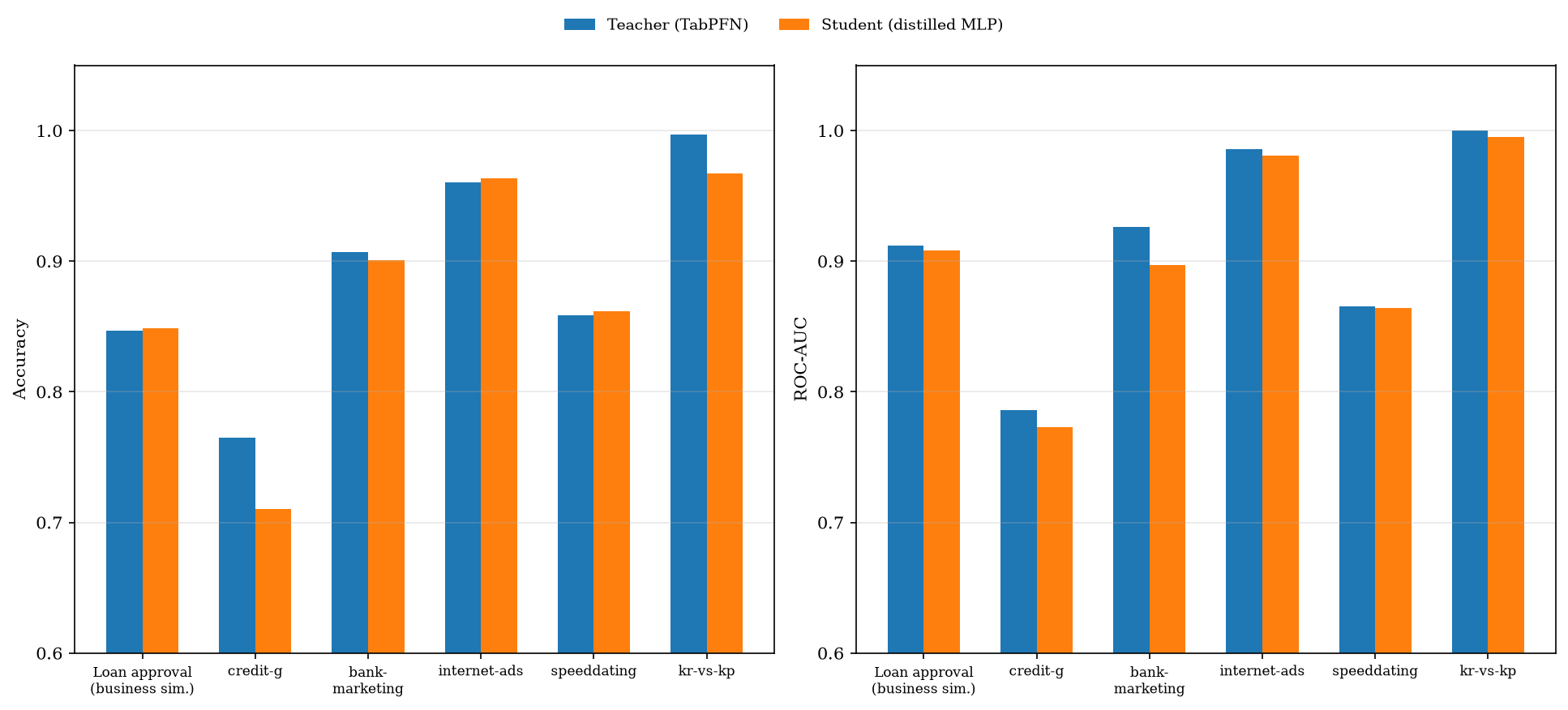}
\caption{Teacher vs. student accuracy (left) and ROC-AUC (right) across six datasets.}
\label{fig:acc-auc}
\end{figure}

\begin{figure}[tbp]
\centering
\includegraphics[width=0.72\textwidth]{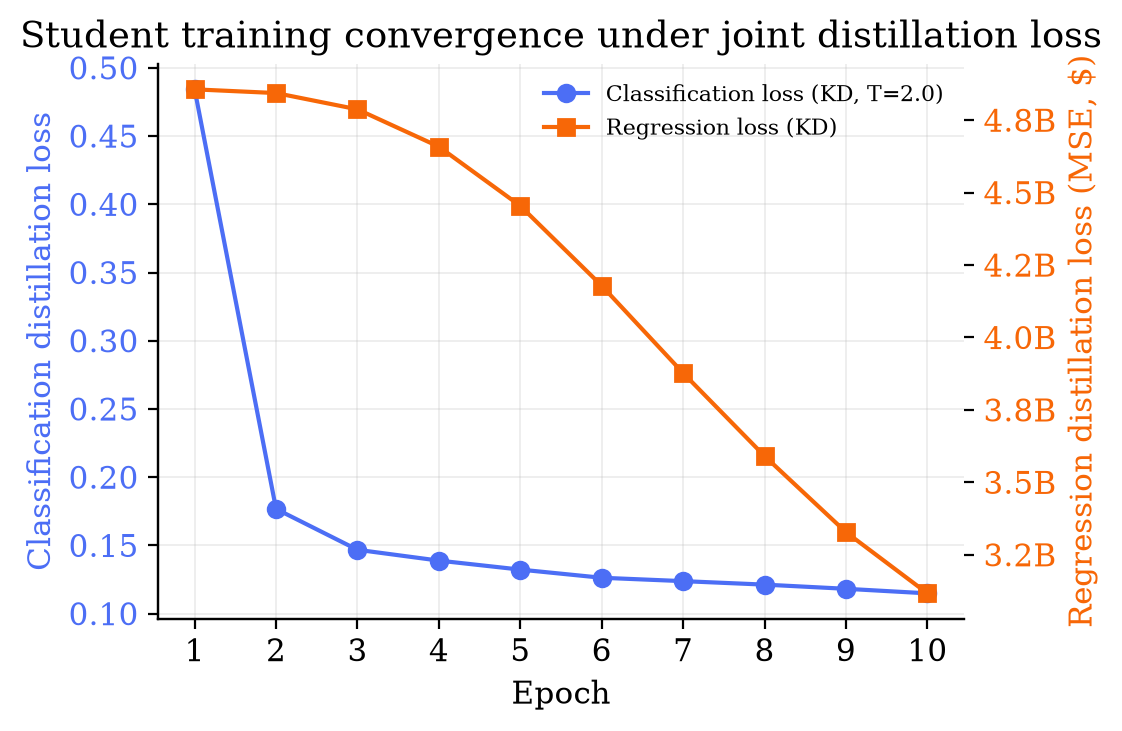}
\caption{Student classification (KD, $T=2.0$) and regression distillation loss over 10 training epochs.}
\label{fig:loss}
\end{figure}

\section{Extended Regression Benchmark Sweeps}
\label{sec:appendix-extended-regression}
\label{sec:regbench}

Section~\ref{sec:m1-regression}'s $\alpha$ control is a single-dataset finding (the primary business-simulation regressor only). To evaluate whether monotonic benefits from teacher soft targets generalize, we ran the identical distillation protocol (\codepath{scripts/run_regression_benchmark.py}: \texttt{TabPFNRegressor} teacher, the same \texttt{SmallRegressor} architecture and loss as Eq.~\ref{eq:kd-reg}, $\alpha=0.9$, $T=1.0$, 200 epochs, seed 42) on five further OpenML regression datasets spanning raw feature counts from 11 to 1,024 and diverse target scales: \textbf{cpu-small} (Computer Activity, did 227) \citep{delve1996cpusmall}, \textbf{cpmp-runtime} (Container Pre-Marshalling Problem algorithm runtime, did 41928) \citep{tierney2015cpmp}, \textbf{topo-2-1} (drug-design topological descriptors, did 422) \citep{openml422topo}, \textbf{coil-20} (Columbia Object Image Library, pixel features $\to$ turntable angle, did 46783) \citep{nene1996coil20}, and \textbf{football-player-position} (player attributes $\to$ position score, did 46763) \citep{heid2025msl}. Unlike the classification datasets of Table~\ref{tab:sources}, all five are used at $\leq$10,000 rows by design (Table~\ref{tab:regbench}), with the two highest-dimensional sets (topo-2-1: 266 features; coil-20: 1,024 features) additionally row-capped below their full size to keep TabPFN's CPU in-context inference tractable.

\begin{table}[tbp]
\centering
\footnotesize
\resizebox{\textwidth}{!}{%
\begin{tabular}{lrrrrrrrr}
\toprule
\textbf{Dataset} & \textbf{$n_{\text{train}}$} & \textbf{$n_{\text{test}}$} & \textbf{Teacher RMSE} & \textbf{Student RMSE} & \textbf{RMSE ret.} & \textbf{Teacher WMAPE} & \textbf{Student WMAPE} & \textbf{WMAPE ret.} \\
\midrule
cpu-small & 6{,}553 & 1{,}639 & 2.4542 & 3.4243 & 71.7\% & 0.0200 & 0.0313 & 63.9\% \\
cpmp-runtime & 1{,}686 & 422 & 431.16 & 526.32 & 81.9\% & 0.1134 & 0.1849 & 61.3\% \\
topo-2-1 & 1{,}600 & 400 & 0.0265 & 0.0263 & 100.0\%\textsuperscript{*} & 0.0204 & 0.0203 & 100.0\%\textsuperscript{*} \\
coil-20 & 1{,}152 & 288 & 0.0357 & 0.1493 & 23.9\% & 0.0017 & 0.0069 & 24.4\% \\
football-player-position & 2{,}888 & 723 & 0.7870 & 0.7932 & 99.2\% & 0.2689 & 0.2879 & 93.4\% \\
\bottomrule
\end{tabular}%
}
\caption{Extended regression benchmark: teacher (TabPFNRegressor) vs.\ distilled student (with teacher signal, $\alpha=0.9$, $T=1.0$, 200 epochs, seed 42; cpu-small and coil-20's student values are 5-fold cross-validated means, cpmp-runtime/topo-2-1/football-player-position a single 80/20 split) on five further OpenML regression datasets. ``Student'' throughout is the distilled (with-teacher-signal) student. Against the hard-label-only ($\alpha=0$) control student (evaluated under an identical protocol), the teacher signal shows a positive WMAPE-retention gap on every one of the five datasets: cpu-small (63.9\% vs.\ 47.2\%, +16.7pp), cpmp-runtime (61.3\% vs.\ 59.0\%, +2.3pp), topo-2-1 (100.0\%\textsuperscript{*} vs.\ 89.7\%, +10.3pp), coil-20 (24.4\% vs.\ 2.5\%, +21.9pp), and football-player-position (93.4\% vs.\ 90.2\%, +3.2pp), exhibiting the same monotonic benefit from soft targets as Section~\ref{sec:m1-regression}'s primary finding. \textsuperscript{*}Retention is defined as $\text{RMSE}_{\text{teacher}} / \text{RMSE}_{\text{student}}$ (capped at 100\% as lower error indicates superior performance); topo-2-1's uncapped ratio is 100.8\% RMSE / 100.4\% WMAPE, confirming parity with the teacher within test-set noise. coil-20 remains the weakest of the five (24.4\% WMAPE retention): a compact 67{,}713-parameter student compressing a 1,024-raw-pixel-feature regression target represents a severe compression challenge for this architecture.}
\label{tab:regbench}
\end{table}

\begin{figure}[tbp]
\centering
\includegraphics[width=0.95\textwidth]{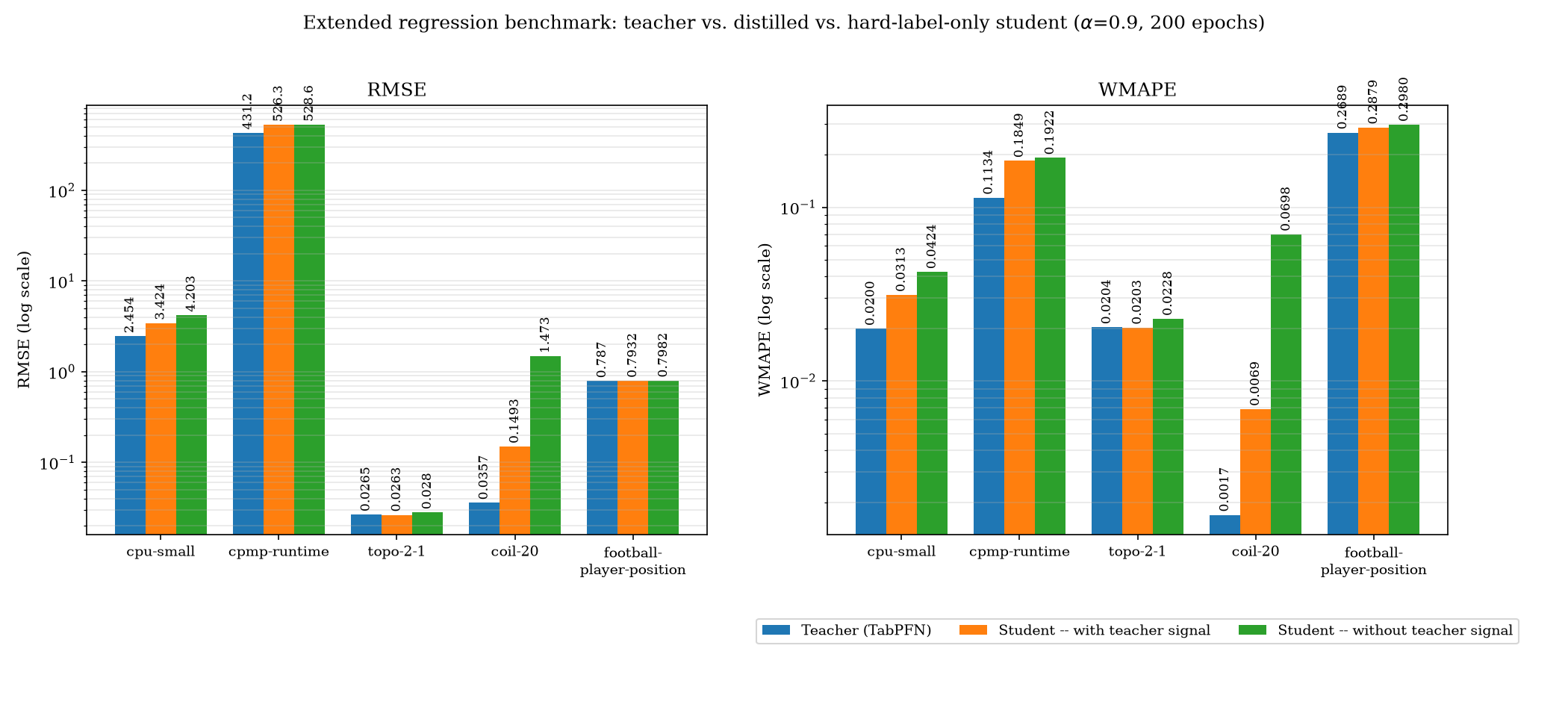}
\caption{Extended regression benchmark (Table~\ref{tab:regbench}): RMSE (left, log scale) and WMAPE (right, log scale) for teacher, distilled student, and hard-label-only student across the five further OpenML regression datasets.}
\label{fig:regbench}
\end{figure}

Two patterns emerge. First, RMSE and WMAPE retention diverge across datasets: cpu-small's WMAPE retention (63.9\%) sits below its RMSE retention (71.7\%) because WMAPE weights proportional error on smaller-magnitude instances more heavily, while cpmp-runtime shows the same direction (61.3\% vs.\ 81.9\%), mirroring the RMSE/WMAPE divergence observed in Section~\ref{sec:m1-regression}. Second, input dimensionality correlates more strongly with student performance than dataset size: topo-2-1 (266 features, 1,600 rows) reaches parity with its teacher, while coil-20 (1,024 features, 1,152 rows) exhibits lower retention (24.4\% WMAPE retention), consistent with capacity limits of the fixed 64--32 hidden layer width for 1,024-dimensional inputs. Evaluated against the hard-label baseline, teacher supervision improves retention across all five datasets, corroborating the monotonic benefit of soft targets identified in Section~\ref{sec:m1-regression}.

\section{Prompt Templates, Schemas, and Parameter Accounting}
\label{sec:appendix-prompts-params}

\subsection{Prompt Templates and Function-Calling Schemas}

\paragraph{Scenario Verbalization Prompt (\codepath{scripts/build_scenario_eval_set.py}).}
The 50-example what-if evaluation set was generated by sampling realistic customer attribute changes and verbalizing them into natural-language questions using \texttt{gpt-4o-mini} with the following system instruction:

\begin{lstlisting}
You write a single short, natural sentence describing a hypothetical change to a loan applicant's profile, for a human loan officer. You MUST explicitly state every new value given to you, exactly as given (same spelling/number), so the sentence is unambiguous. Do not mention JSON, field names verbatim, or add any values not given.
\end{lstlisting}

\paragraph{Cloud LLM Orchestrator System Prompt (\codepath{scripts/hybrid_architecture.py}).}
The cloud orchestrator (\texttt{gpt-4o-mini}) performs tool selection and scenario parsing via the following system prompt:

\begin{lstlisting}
You are a loan analysis tool orchestrator.

Your job: 
1. Decide which tool to call
2. Extract ALL parameters including scenario modifications
3. For whatif queries, parse the natural language scenario into structured fields

When user asks "what if age is 35 and education is Masters":
- Call: load_predict_loan_approval_whatif
- Extract: customer_id AND modifications={"age": 35, "education": "Masters"}

Extract ALL mentioned fields accurately. Valid fields:
- age (number)
- education (Doctorate, Masters, Bachelors, Some-college, HS-grad, 11th, etc.)
- education-num (1-16)
- workclass (Private, Self-emp-inc, Federal-gov, etc.)
- occupation (Tech-support, Sales, Exec-managerial, etc.)
- marital-status (Married-civ-spouse, Divorced, Never-married, etc.)
- relationship (Wife, Husband, Own-child, etc.)
- race (White, Black, Asian-Pac-Islander, Amer-Indian-Eskimo, Other)
- sex (Male, Female)
- capital-gain (number)
- capital-loss (number)
- hours-per-week (number)
- native-country (United-States, Mexico, etc.)

DO NOT call tools without required parameters.
\end{lstlisting}

\paragraph{OpenAI Function Calling Schemas.}
The tool interfaces are strictly constrained using JSON schemas bound to \texttt{gpt-4o-mini}. For what-if scenario prediction, the schema declares parameter extraction explicitly:

\begin{lstlisting}
{
  "type": "function",
  "function": {
    "name": "load_predict_loan_approval_whatif",
    "description": "Predict loan approval for a what-if scenario with feature modifications",
    "parameters": {
      "type": "object",
      "properties": {
        "customer_id": {
          "type": "integer",
          "description": "Customer ID"
        },
        "modifications": {
          "type": "object",
          "description": "Structured feature modifications extracted from scenario. Keys are feature names (age, education, workclass, etc.), values are new values.",
          "additionalProperties": true
        }
      },
      "required": ["customer_id"]
    }
  }
}
\end{lstlisting}

\paragraph{Local SLM Response Formatter Prompt (\codepath{scripts/response_formatter.py}).}
On-premises response generation is executed by \texttt{Qwen2.5-3B-Instruct} without cloud API egress, converting raw tool output and the user query into a clean JSON response:

\begin{lstlisting}
You are a response formatter for a loan analysis system.

Your job: Convert raw tool output and user query into a clean, natural JSON response.

RULES:
1. Return ONLY valid JSON - no explanations, no tool mentions
2. Structure the response to directly answer the user's question
3. Use clear, business-friendly language
4. Include all relevant data from the tool output
5. Do NOT mention "tool", "execution", or technical details

EXAMPLES:

User Query: "Get information for customer 123"
Tool Output: [{"customer_id": 123, "age": 45, "education": "Bachelors", "loan_status": 1}]
Response: {
  "customer_id": 123,
  "age": 45,
  "education": "Bachelors",
  "current_loan_status": "Approved"
}

User Query: "What if customer 456 has a masters degree?"
Tool Output: {"customer_id": 456, "modifications_applied": {"education": "Masters"}, "baseline_loan_approval_probability": 0.32, "modified_loan_approval_probability": 0.78, "probability_change": "+46.0%"}
Response: {
  "customer_id": 456,
  "scenario": "Education changed to Masters",
  "baseline_approval_probability": "32%",
  "new_approval_probability": "78%",
  "change": "+46%",
  "recommendation": "Significant improvement in approval probability"
}

User Query: "What is the maximum loan amount for customer 100?"
Tool Output: {"customer_id": 100, "max_loan_amount": 150000.0}
Response: {
  "customer_id": 100,
  "maximum_loan_amount": "$150,000"
}

Now format the response.
\end{lstlisting}

\subsection{Observability Instrumentation and Trace Schema}

Table~\ref{tab:observability} details the complete OpenTelemetry-compatible span instrumentation for both the hybrid and full-LLM agentic pipelines managed via LangSmith.

\begin{table}[tbp]
\centering
\small
\resizebox{\textwidth}{!}{%
\begin{tabular}{p{4.1cm}p{6.3cm}p{2.3cm}}
\toprule
\textbf{Span} & \textbf{Captures} & \textbf{Instrumentation} \\
\midrule
\texttt{hybrid\_agent.chat} / \texttt{full\_llm\_agent.chat} & Full turn: query, tool selected, extracted args, final answer, end-to-end latency & Explicit (\texttt{@traceable}) \\
Orchestrator LLM call (\texttt{gpt-4o-mini}) & Prompt, completion, token usage, cost & Automatic (LangChain runnable) \\
\texttt{tabpfn\_teacher.predict\_\{approval,amount\}} & Input row, predicted probability/amount, call latency & Explicit \\
\texttt{student\_mlp.predict\_\{approval,amount\}} & Input row, predicted probability/amount, call latency & Explicit \\
\texttt{qwen3b\_formatter} & Prompt, raw tool JSON, formatted answer, latency & Explicit \\
Eval-time feedback (tool-call accuracy, parsing recall/precision, faithfulness, context/answer relevance) & Attached post-hoc onto the matching run via \mbox{\texttt{Client.create\_feedback}} & Explicit, from \codepath{scripts/run_hybrid_eval.py} \\
\bottomrule
\end{tabular}%
}
\caption{What each traced span captures in the LangSmith-instrumented pipeline, and whether the capture is automatic (a consequence of using LangChain runnables) or required an explicit \texttt{@traceable} annotation.}
\label{tab:observability}
\end{table}

\subsection{Detailed PyTorch Layer-by-Layer Parameter Accounting}

Table~\ref{tab:layer-params} reports the exact layer-by-layer parameter accounting for the student MLP architectures (\texttt{SmallClassifier} and \texttt{SmallRegressor}) evaluated on the primary business-decision dataset ($d_{\text{in}}=99$). Both architectures use identical hidden layer dimensions ($64$ and $32$) with Dropout ($p=0.2$) and differ only in output dimension and activation functions.

\begin{table}[tbp]
\centering
\small
\resizebox{\textwidth}{!}{%
\begin{tabular}{lllrr}
\toprule
\textbf{Model} & \textbf{Layer} & \textbf{Shape} & \textbf{Weights + Biases} & \textbf{Total Params} \\
\midrule
\multirow{4}{*}{\textbf{SmallClassifier}} 
  & \texttt{fc1} (Linear + ReLU + Dropout) & $99 \to 64$ & $(64 \times 99) + 64$ & 6{,}400 \\
  & \texttt{fc2} (Linear + Sigmoid + Dropout) & $64 \to 32$ & $(32 \times 64) + 32$ & 2{,}080 \\
  & \texttt{fc3} (Linear, 2-class logits) & $32 \to 2$ & $(2 \times 32) + 2$ & 66 \\
  \cmidrule{2-5}
  & \multicolumn{3}{l}{\textbf{Total Classifier Parameters}} & \textbf{8{,}546} \\
\midrule
\multirow{4}{*}{\textbf{SmallRegressor}} 
  & \texttt{fc1} (Linear + ReLU + Dropout) & $99 \to 64$ & $(64 \times 99) + 64$ & 6{,}400 \\
  & \texttt{fc2} (Linear + ReLU + Dropout) & $64 \to 32$ & $(32 \times 64) + 32$ & 2{,}080 \\
  & \texttt{fc3} (Linear, continuous scalar) & $32 \to 1$ & $(1 \times 32) + 1$ & 33 \\
  \cmidrule{2-5}
  & \multicolumn{3}{l}{\textbf{Total Regressor Parameters}} & \textbf{8{,}513} \\
\midrule
\multicolumn{4}{l}{\textbf{Combined Deployed Pipeline Parameters (Classifier + Regressor)}} & \textbf{17{,}059} \\
\bottomrule
\end{tabular}%
}
\caption{Layer-by-layer parameter breakdown for the feed-forward student networks on the primary loan dataset ($d_{\text{in}}=99$ one-hot encoded features). Parameter counts match the values reported in Table~\ref{tab:params} exactly.}
\label{tab:layer-params}
\end{table}

\section{Reproducibility and Environment Details}
\label{sec:appendix-reproducibility}

\subsection{Code, Data, and Artefacts}
Code, data, and evaluation artefacts are available at \url{https://github.com/nitsourish/Model-Distillation-Agentic-Workflow}. All numbers in this paper are computed directly from artefacts checked into that repository: \codepath{artefacts/distillation_metrics.json}, \codepath{artefacts/hybrid_eval_50_results.json}, \codepath{artefacts/hybrid_eval_50_results_rtx4060ti.json}, \codepath{artefacts/corrected_rerun/hybrid_eval_50_results_rerun.json}, \codepath{artefacts/parse_accuracy_eval_results*.json}, and parameter counts measured from \codepath{artefacts/student_classifier.pth}, \codepath{artefacts/student_regressor.pth} and a live-loaded \texttt{tabpfn==8.2.0} model. The extended CC18 and regression-benchmark datasets (Appendices~\ref{sec:appendix-extended-classification} and \ref{sec:appendix-extended-regression}) are reproducible via \codepath{scripts/run_classification_benchmark.py}, \codepath{scripts/run_regression_benchmark.py}, and \codepath{scripts/download_openml_cc18.py} against each dataset's public OpenML source; their raw result artefacts and source CSVs are not checked in, since the paper's own tables are the recorded snapshot. 

\subsection{Environment and Dependencies}
The software environment is managed using \texttt{uv} with Python $\ge 3.11$. Key library versions include:
\begin{itemize}
\item \texttt{torch>=2.13.0} and \texttt{transformers>=4.57.0} (student training and local SLM inference)
\item \texttt{tabpfn==8.2.0} (Prior Labs TabPFN foundation model)
\item \texttt{langchain-openai>=1.0.1} and \texttt{langgraph>=1.0.2} (agentic orchestration)
\item \texttt{langsmith>=0.10.13} (observability tracing and evaluation feedback)
\item \texttt{fairlearn>=0.13.0} (subgroup fairness metrics)
\item \texttt{openml>=0.15.1} and \texttt{scikit-learn>=1.9.0} (dataset fetching, preprocessing, and controls)
\item \texttt{outlines>=0.1.11} (constrained JSON generation for SLM ablation)
\end{itemize}

\subsection{Hardware Profiles and Extended Latency Measurements}
Three hardware configurations were evaluated. The first two are shared development machines; the third is a dedicated check. They are not interchangeable ``CPU'' measurements.
\begin{enumerate}
\item \textbf{Shared CPU host}: 32GB RAM x86\_64, PyTorch CPU execution (subject to OS contention).
\item \textbf{Shared consumer GPU}: NVIDIA GeForce RTX 4060 Ti 16GB VRAM, CUDA 12.x; \texttt{bfloat16} for local Qwen; TabPFN teacher also on GPU in the full-LLM columns of Table~\ref{tab:costlat}.
\item \textbf{Dedicated M5 Pro CPU}: Apple M5 Pro, 48GB unified RAM, CPU backend only. Absolute latencies are lower than item~1 because the host is unloaded and the SoC differs, not because the protocol changed.
\end{enumerate}

\begin{figure}[tbp]
\centering
\includegraphics[width=0.68\textwidth]{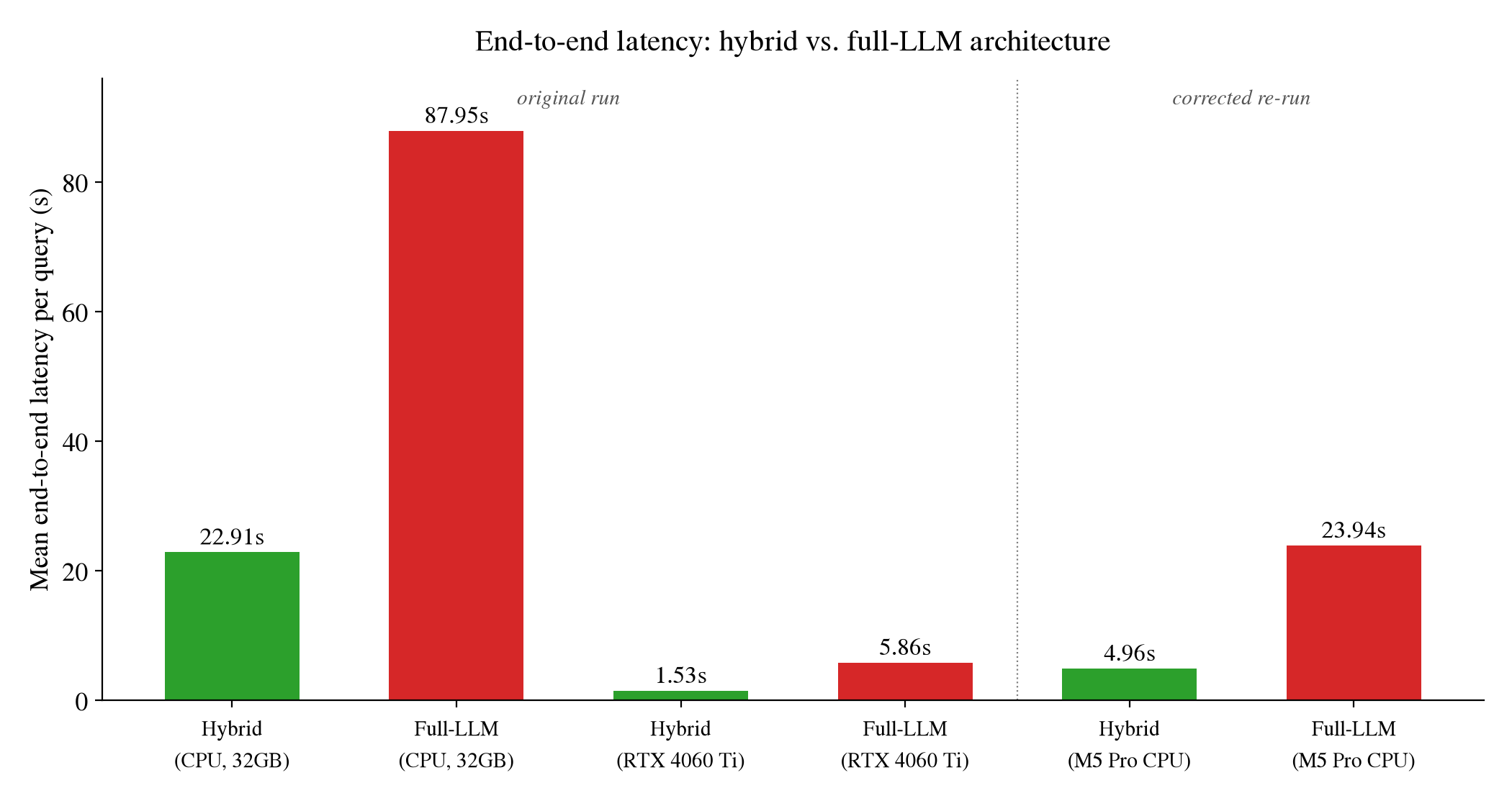}
\caption{End-to-end mean query latency, hybrid vs.\ full-LLM, on three machines: shared 32GB CPU host, shared RTX 4060 Ti (Table~\ref{tab:costlat}), and dedicated Apple M5 Pro CPU (Table~\ref{tab:costlatm5}). Ordering and $\sim$4$\times$ ratio hold; absolute seconds on the two CPU machines are not the same experiment.}
\label{fig:latency}
\end{figure}

\begin{figure}[tbp]
\centering
\includegraphics[width=0.58\textwidth]{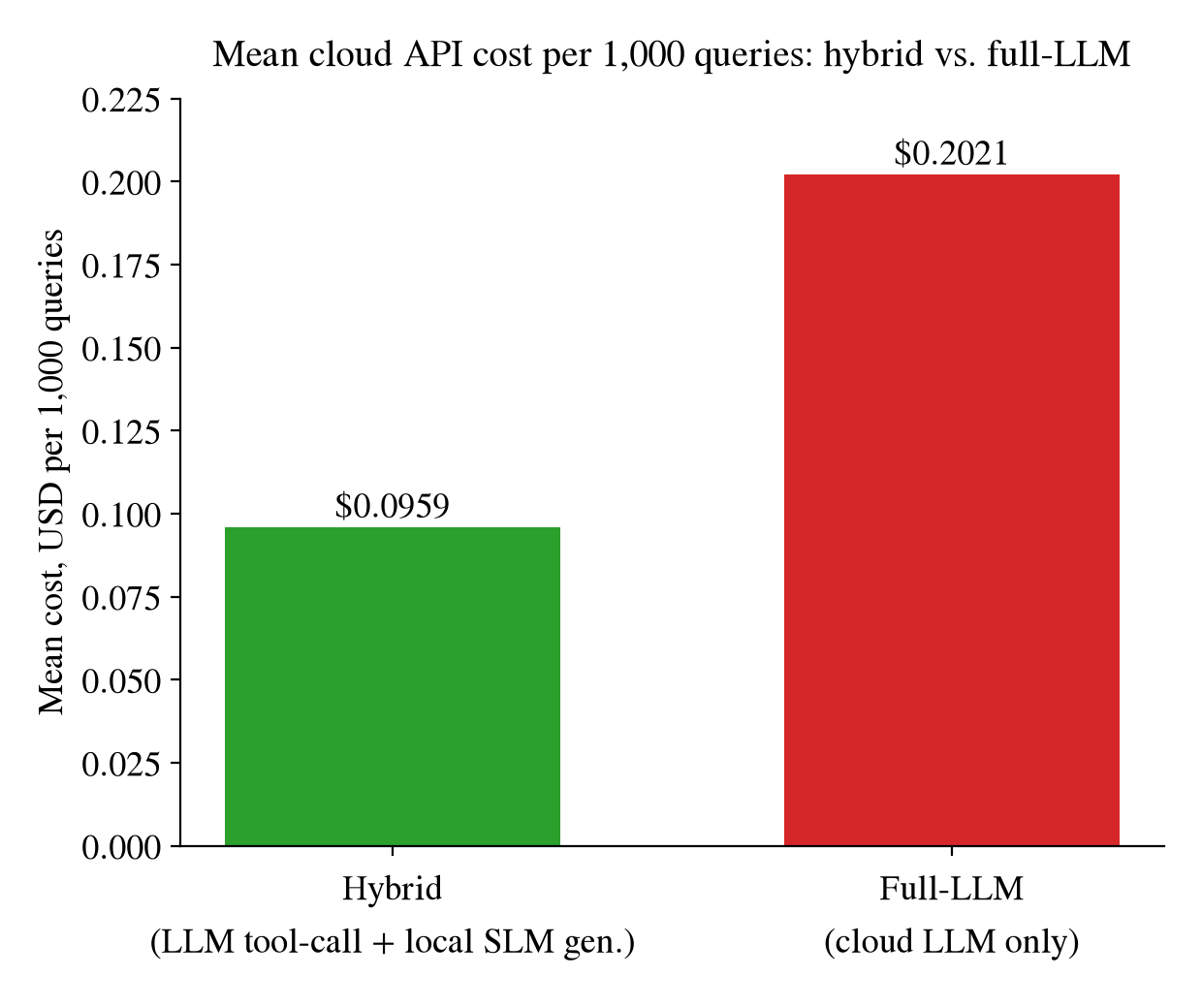}
\caption{Mean cloud API cost per 1,000 queries, hybrid vs.\ full-LLM. Token spend does not depend on local CPU/GPU; M5 Pro dollars in Table~\ref{tab:costlatm5} differ only by a separate API pass.}
\label{fig:cost}
\end{figure}

\begin{figure}[tbp]
\centering
\includegraphics[width=0.68\textwidth]{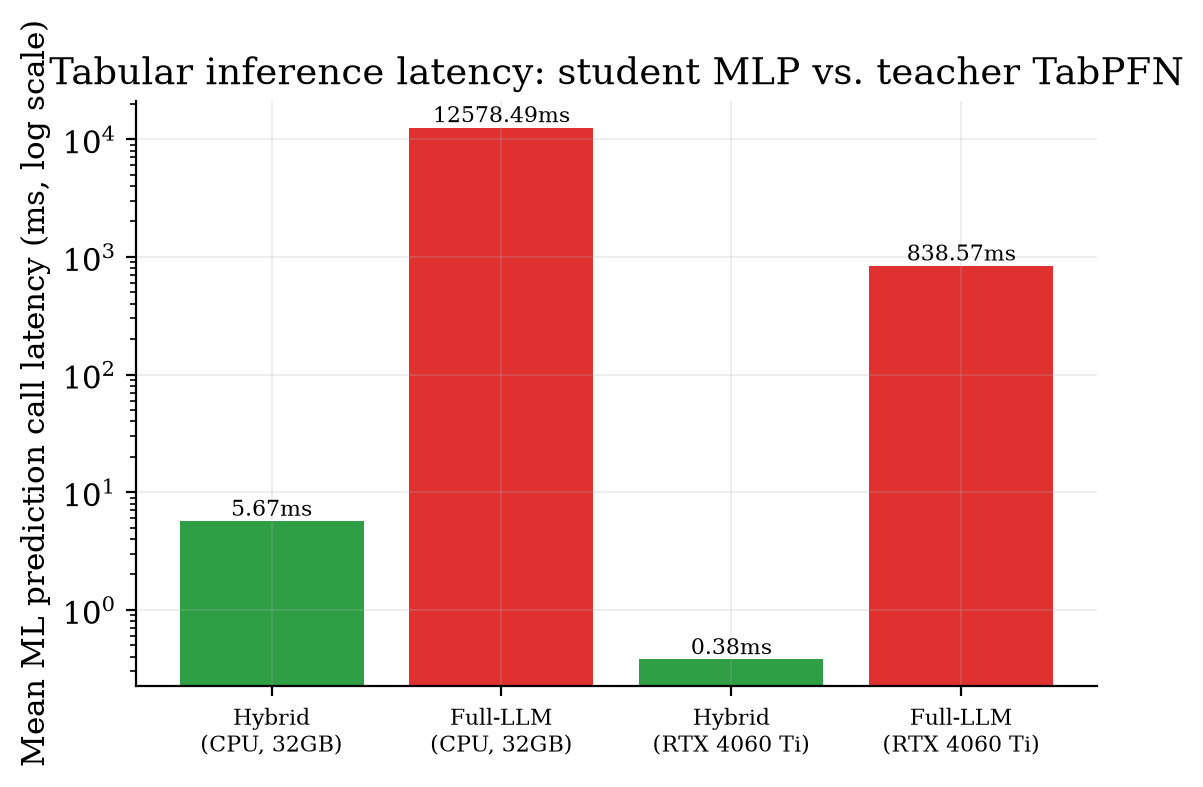}
\caption{Tabular-model inference latency only (student MLP vs. tabular foundation-model teacher), log scale.}
\label{fig:predict}
\end{figure}

Table~\ref{tab:costlatm5} reports the detailed performance profile measured on the Apple M5 Pro CPU. Figure~\ref{fig:latency} plots total query latency across all three hardware profiles, Figure~\ref{fig:cost} visualizes the per-query cloud API cost comparison, and Figure~\ref{fig:predict} plots the tabular prediction latency on a log scale.

\begin{table}[tbp]
\centering
\footnotesize
\begin{tabular}{lrr}
\toprule
 & \textbf{Hybrid} & \textbf{Full-LLM} \\
\midrule
Mean cost / query & \multicolumn{2}{c}{\$9.51$\times10^{-5}$ / \$2.04$\times10^{-4}$} \\
Mean orchestrator latency & 0.86s & 0.89s \\
Mean formatter/gen. latency & 4.10s & 1.51s \\
Mean ML predict latency & 6.09ms & 19{,}579ms \\
Unaccounted (tool-exec.\ wall time) & $\approx$0.01s & 1.97s \\
\midrule
Mean total latency (measured) & 4.96s & 23.94s \\
\bottomrule
\end{tabular}
\caption{Cost and latency, 50-query benchmark on a \emph{dedicated} Apple M5 Pro (CPU backend, 48GB). Measured independently from Table~\ref{tab:costlat}; the small cost difference (\$9.51$\times10^{-5}$ / \$2.04$\times10^{-4}$ vs.\ \$9.59$\times10^{-5}$ / \$2.02$\times10^{-4}$) is token-count noise on a new \texttt{gpt-4o-mini} pass, not a hardware effect. Hybrid-vs-full-LLM ratios remain similar (4.82$\times$ latency, 2.15$\times$ cost). Orchestrator times here (0.86s / 0.89s) match the GPU profile, supporting the claim that 12.75s on the shared CPU host is contention. The full-LLM residual shrinks from 67.88s to 1.97s.}
\label{tab:costlatm5}
\end{table}

\subsection{Standalone SLM-as-Parser Ablation}
\label{sec:slm-ablation}

To evaluate whether the cloud LLM step can itself be eliminated by having the local SLM perform scenario parsing directly, closing the loop entirely on-premises, we ran a standalone 50-example benchmark (\codepath{scripts/run_parse_accuracy_eval.py}) comparing \texttt{gpt-4o-mini} against base \texttt{Qwen2.5-3B-Instruct} (without task-specific fine-tuning) executing constrained JSON decoding via Outlines in a zero-shot setting. Table~\ref{tab:slmablation} reports two independent runs.

\begin{table}[tbp]
\centering
\small
\begin{tabular}{lrrrr}
\toprule
\textbf{Configuration} & \textbf{Tool-call acc.} & \textbf{Mean F1} & \textbf{Mean cycle latency} & \textbf{Cost / call} \\
\midrule
Cloud LLM (gpt-4o-mini) & 0.38 & 0.565 & 28.18s & \$2.79$\times10^{-4}$ \\
Local SLM, zero-shot (run 1) & 0.58 & 0.801 & 2.13s & \$0 \\
Local SLM, zero-shot (run 2) & 0.58 & 0.776 & 2.03s & \$0 \\
\bottomrule
\end{tabular}
\caption{Standalone SLM-as-parser ablation, zero-shot only. Exact-match tool-call accuracy on extracted modification sets; mean F1 is the harmonic mean of precision/recall.}
\label{tab:slmablation}
\end{table}

The zero-shot 3B SLM achieves 0.58 exact-match accuracy consistently across both runs. While higher than \texttt{gpt-4o-mini}'s 0.38 under this identical, generic schema (which penalizes the cloud model by omitting named parameter declarations), it remains far below the 98\% accuracy achieved by the hybrid pipeline using genuine schema-constrained function calling (Table~\ref{tab:toolcall}). Small open models are not reliably instruction-tuned for OpenAI-style function calling, making constrained JSON decoding via Outlines their best available local interface. Qualitative failure analysis shows that the small model frequently copies unmentioned fields from the applicant context into the modification JSON, failing to isolate user-specified diffs. This empirically justifies why the production hybrid architecture retains the cloud LLM for tool selection and scenario parsing, while offloading numeric prediction and natural-language formatting to local modules.

\section{Extended Discussion, Subgroup Fairness, and Threats to Validity}
\label{sec:appendix-validity}

\subsection{Detailed Subgroup Fairness Protocol}
As noted in Section~\ref{sec:discussion}, soft-target compression is not guaranteed to preserve a teacher's group-conditional error rates even when aggregate accuracy/AUC is maintained. To audit fairness, we evaluated demographic parity difference (max-min predicted positive rate) and equalized-odds difference (max-min of true-positive and false-positive rates) across protected groups on the UCI Adult test set using Fairlearn \citep{fairlearn2020}. 

For sex, the groups comprised 679 female and 1,321 male applicants. Demographic parity difference was $0.180$ (teacher) vs. $0.176$ (distilled student); equalized odds difference was $0.093$ (teacher) vs. $0.080$ (student). For race, evaluating four groups with $\ge 20$ test rows (White, Black, Asian-Pac-Islander, Other), demographic parity difference widened from $0.240$ (teacher) to $0.256$ (student), and equalized odds difference widened from $0.431$ to $0.490$ (+14\% relative). While true-positive rates increased across all racial groups under student predictions, the non-uniform rate of change across groups widened the disparate-impact gap. This demonstrates empirically why aggregate accuracy retention alone cannot validate compression safety in sensitive domains.

\subsection{Detailed Accounting of Threats to Validity}
The six items in Section~\ref{sec:limitations} are restated here with additional operational detail.
\begin{enumerate}
\item \textbf{Orchestrator Tier Representation}: \texttt{gpt-4o-mini} served as an accessible proxy for the target \texttt{gpt-4o-nano} production tier. While functional capabilities and schema adherence are expected to align, absolute token pricing and inference latency will differ when deployed against smaller nano-tier APIs.
\item \textbf{Evaluation Benchmark Size}: The 50-query what-if evaluation set establishes clear, statistically significant separation in latency, tool-call accuracy, and parameter accuracy. However, small sample counts constrain statistical power when evaluating subtle natural-language variations across LLM judges.
\item \textbf{Synthetic Regression Label}: The regression target \texttt{max\_loan} was generated via a synthetic heuristic simulation for what-if exploration. While suitable for evaluating continuous distillation dynamics, it serves purely as an experimental simulation target.
\item \textbf{Hardware Concurrency and Execution Environments}: Sweeps across the 32GB CPU host and RTX 4060 Ti GPU were collected on shared development machines subject to transient OS contention. Orchestrator wall-clock of 12.75s on that CPU host vs.\ $\approx$0.85s on GPU/M5 is the same \texttt{gpt-4o-mini} API under load. The clean-room reproduction on the Apple M5 Pro host verified that the measured speedup \emph{ratio} is robust; absolute CPU seconds are not.
\item \textbf{Single-Seed Protocols}: Primary training and evaluation sweeps were executed with fixed seed 42 to enable direct paired comparisons. Test-set variance was accounted for via bootstrap confidence intervals, but multi-seed weight-initialization sweeps remain future work.
\item \textbf{Fairness Audit Scope}: Subgroup evaluations were focused on the two well-represented protected attributes in the Adult dataset (sex and race). Attributes with small sample sizes were excluded to prevent deceptive small-sample statistics.
\end{enumerate}

\end{document}